\RequirePackage{fix-cm}
\documentclass[twocolumn,compsoc]{CVM}

\graphicspath{{figures/},{figures/cvm/},{figures/Imgs/},{figures/pictures/}}

\usepackage{amsmath}
\usepackage{graphicx}
\usepackage{multirow}
\usepackage{subfigure}
\usepackage{caption}
\usepackage{rotating}
\usepackage{cite}
\usepackage{amssymb}
\usepackage{array}
\usepackage{bm}
\usepackage{color}
\usepackage{enumitem}
\usepackage{overpic}
\usepackage{colortbl}
 
\let\Letter\relax
\usepackage{marvosym}
\usepackage{pifont}

\definecolor{green}{RGB}{0,139,69} 
\usepackage[colorlinks,
            linkcolor=blue,
            anchorcolor=blue,
            citecolor=blue]{hyperref}

\usepackage{hyperref}
\makeatletter
\def\UrlAlphabet{%
      \do\a\do\b\do\c\do\d\do\e\do\f\do\g\do\h\do\i\do\j%
      \do\k\do\l\do\m\do\n\do\o\do\p\do\q\do\r\do\s\do\t%
      \do\u\do\v\do\w\do\x\do\y\do\z\do\A\do\B\do\C\do\D%
      \do\E\do\F\do\G\do\H\do\I\do\J\do\K\do\L\do\M\do\N%
      \do\O\do\P\do\Q\do\R\do\S\do\T\do\U\do\V\do\W\do\X%
      \do\Y\do\Z}
\def\UrlDigits{\do\1\do\2\do\3\do\4\do\5\do\6\do\7\do\8\do\9\do\0}
\g@addto@macro{\UrlBreaks}{\UrlOrds}
\g@addto@macro{\UrlBreaks}{\UrlAlphabet}
\g@addto@macro{\UrlBreaks}{\UrlDigits}
\makeatother

\definecolor{mygray}{gray}{.90}

\newcommand{\ie}{\textit{i}.\textit{e}.}
\newcommand{\eg}{\textit{e}.\textit{g}.}

\def\ourmodel{\textit{FSNet}}

\def\ManuscriptInfo{Manuscript received: 2021-08-31; accepted: 20xx-xx-xx.}

\def\PaperTitle{Full-Duplex Strategy for Video Object Segmentation}

\def\AuthorNames{
Ge-Peng~Ji$^{1}$, Deng-Ping Fan$^{*}$(\Letter), Keren Fu$^2$, Zhe Wu$^3$, Jianbing Shen$^4$, and Ling Shao$^5$}

\def\AuthorAdress{
* & College of Computer Science, Nankai University, Tianjin, China.\\
$1\quad$ & School of Computer Science, Wuhan University, Wuhan, China.\\
$2\quad$ & College of Computer Science, Sichuan University, China.\\ 
$3\quad$ & Peng Cheng Laboratory, Shenzhen, China.\\
$4\quad$ & School of Computer Science, Beijing Institute of Technology, Beijing, China. \\
$5\quad$ & IIAI, Abu Dhabi, UAE.\\ 
&Corresponding author: dengpingfan@mail.nankai.edu.cn\\
}

\def\firstheader{DOI 10.1007/s41095-xxx-xxxx-x\hfill Vol. x, No. x, month year, xx--xx\\[5mm]~Research Article}

\headevenname{{G.-P. Ji, D.-P. Fan, K. Fu, and Z. Wu, \etal} }

\begin{document}

\MakePageStyle

\MakeAbstract{
Previous video object segmentation approaches mainly focus on using simplex solutions between appearance and motion, limiting feature collaboration efficiency among and across these two cues.
In this work, we study a novel and efficient full-duplex strategy network (\textbf{\ourmodel}) to address this issue, by considering a better mutual restraint scheme between motion and appearance in exploiting the cross-modal features from the fusion and decoding stage. Specifically, we introduce the 
relational cross-attention module (RCAM) to achieve bidirectional message propagation across embedding sub-spaces. To improve the model's robustness and update the inconsistent features from the spatial-temporal embeddings, we adopt the bidirectional purification module (BPM) after the RCAM.  
Extensive experiments on five popular benchmarks show that our \ourmodel~is 
robust to various challenging scenarios (\eg, motion blur, occlusion) and 
achieves favourable performance against existing cutting-edges both in the video object segmentation and video salient object detection tasks.
The project is publicly available at: 
\url{https://dpfan.net/FSNet}.

}

\MakeKeywords{Video Object Segmentation, Salient Object Detection, Visual Attention}

\section{Introduction}\label{sec:introduction}

\begin{figure}[t!]
  \centering
  \includegraphics[width=.99\columnwidth]{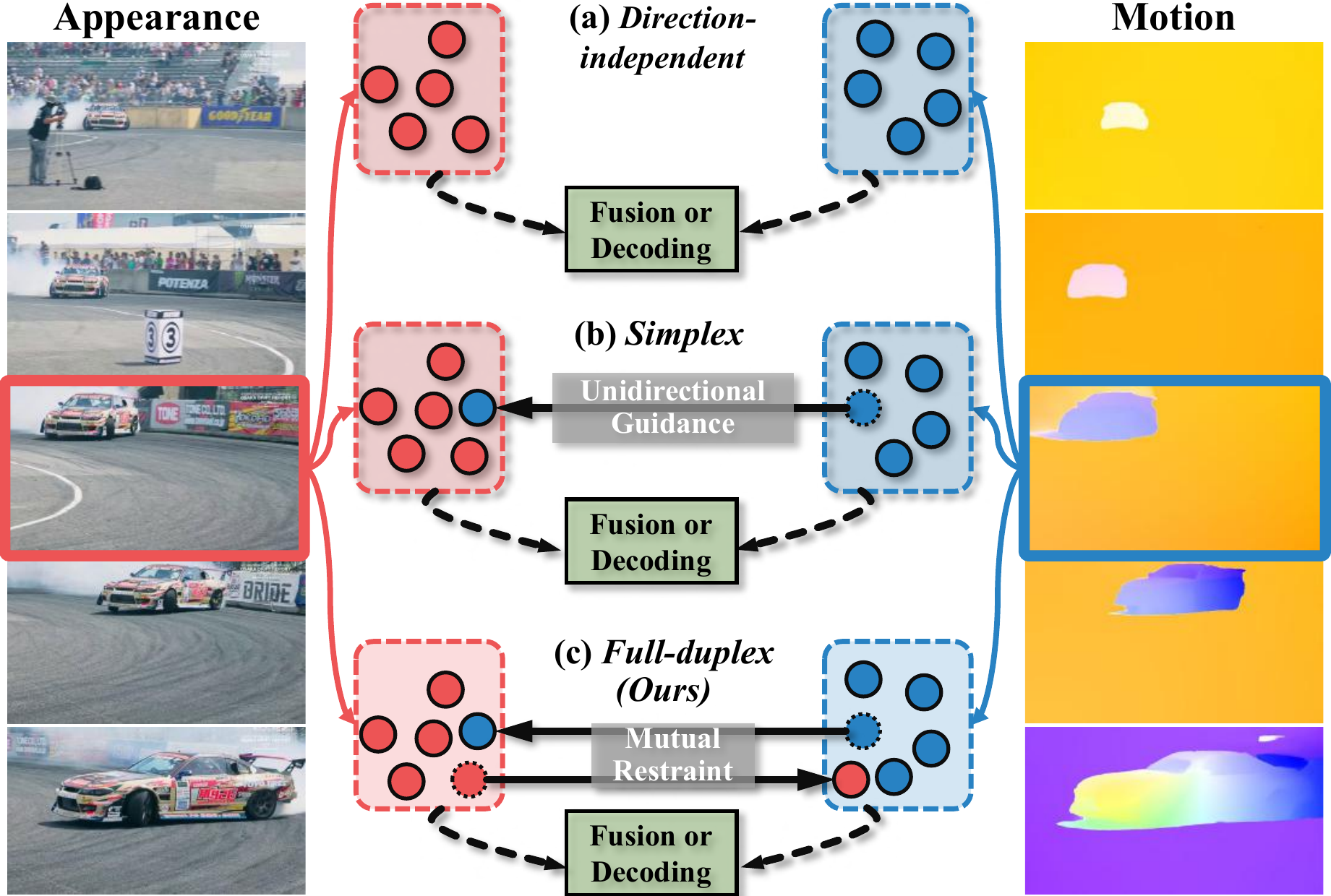}
  \caption{
  Comparison between three strategies for embedding appearance and motion patterns before the fusion and decoding stage.
  (a) \textit{Direction-independent} strategy~\cite{jain2017fusionseg} without information transmission, 
  (b) \textit{Simplex} strategy~\cite{zhou2020motion_attentive} with only unidirectional information transmission, \eg, using motion guides appearance or vice versa, and 
  (c) our \textit{full-duplex} strategy 
  with simultaneously bidirectional information transmission. 
  This paper mainly focuses on discussing directional modelling (b \& c) in the deep learning era.
  }\label{fig:teaser_figure}
\end{figure}

Over the past three years, social platforms have accumulated a large number of short videos. Analyzing these videos efficiently and intelligently has become a challenging issue today. 
Video object segmentation (VOS)~\cite{wang2021end,chen2020state,wen2020DMVOS,huang2020fast} is a fundamental technique to address this issue, whose purpose is to delineate pixel-level moving object\footnote{We use `foreground object' \& `target object' interchangeably.} masks in each frame.
Besides video analysis, many other applications have also benefited from VOS, such as  robotic manipulation~\cite{abramov2012depth}, autonomous cars~\cite{maddern20171}, video editing~\cite{jain2016click}, action segmentation~\cite{wang2020context}, optical flow estimation~\cite{ding2020every}, medical diagnosis~\cite{ji2021pnsnet}, interactive segmentation~\cite{chen2020scribblebox,miao2020memory,heo2020interactive,yin2021learning,cheng2021modular}, URVOS~\cite{seourvos}, and video captioning~\cite{pan2017video}.

Recently, we have witnessed rapid development in addressing video object understanding by exploiting the relationships between the frames' appearance-aware~\cite{zelnik2001event,chang2013video,lee2017contour,reso2013temporally,van2013online} and motion-aware~\cite{ilg2017flownet,teed2020raft,shi1994good}.
Unfortunately, short-term dependency prediction~\cite{ilg2017flownet,teed2020raft} generates unreliable estimations and suffers the common ordeals~\cite{hu2020motion} (\eg, noise, deformation, and diffusion). In addition, the capability of appearance-based modelling like recurrent neural network~\cite{tokmakov2017learning1,bmvc2020making,fan2019shifting} (RNN) is severely hindered by blurred foregrounds or cluttered backgrounds~\cite{chen2019motion}.
Those conflicts are prone to accumulating inaccuracies and the propagation of spatial-temporal embeddings, which cause the short-term feature drifting problem~\cite{yang2019anchor}.

As shown in \figref{fig:teaser_figure}~(a), the \textit{direction-independent strategy}~\cite{khoreva2017lucid,jain2017fusionseg,cheng2017segflow,tokmakov2017learning1,xiao2019online} is the earliest solution by encoding the appearance and motion features individually and fuse them directly.
However, this intuitive way will implicitly cause feature conflicts since the motion- and appearance-aware features are derived from two distinctive modalities, which is extracted from separate branches.
An alternative way is to integrate them in a guided manner. As illustrated in ~\figref{fig:teaser_figure}~(b), several recent methods opt for the \textit{simplex strategy}~\cite{zhou2020motion_attentive,tsai2016video,lin2020flow,nilsson2018semantic,li2019motion,hu2020motion,peng2019automatic}, which is either appearance-based or motion-guided. 
Although these two strategies have achieved promising results, they both fail to consider the \textbf{mutual restraint} between the appearance and motion features that both guide human visual attention allocation during dynamic observation, according to previous studies in cognitive psychology~\cite{koch1987shifts,wolfe1989guided,treisman1980feature} and computer vision~\cite{wang2020paying,jain2017fusionseg}. 

\begin{figure}[t!]
  \centering
  \includegraphics[width=\columnwidth]{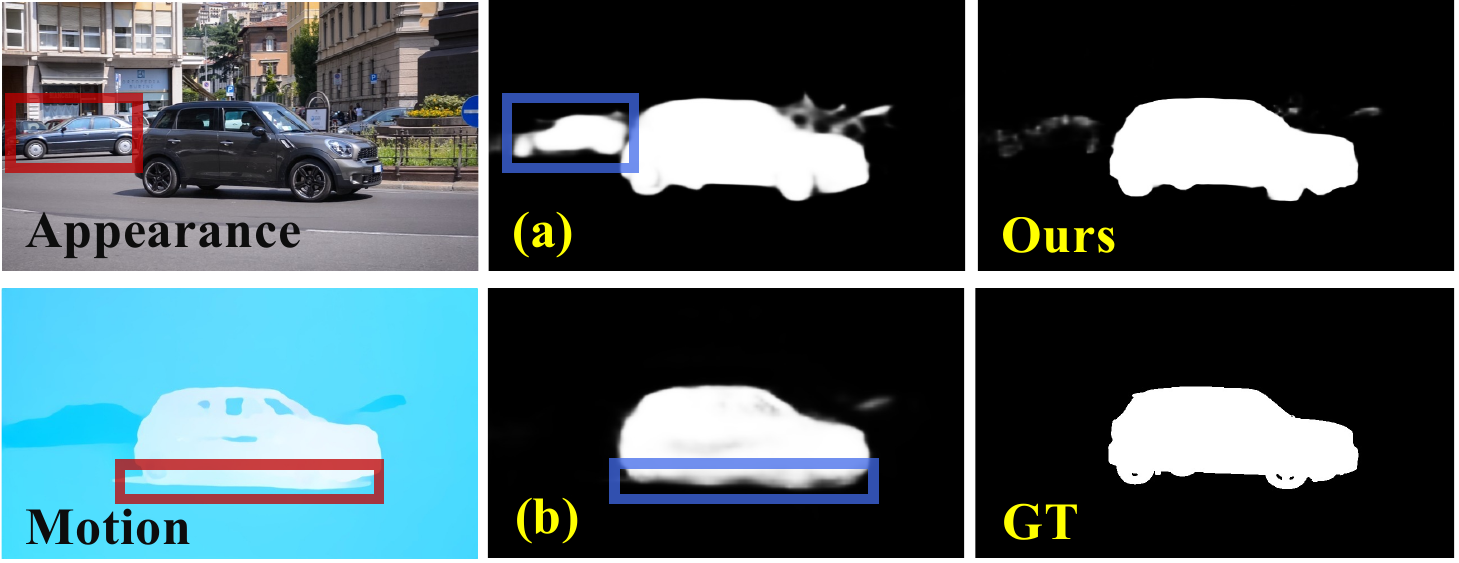}
  \caption{
  Visual comparison between the \textit{simplex} (\ie, (a) appearance-refined motion and (b) motion-refined appearance) and our \textbf{\textit{full-duplex} strategy} under our framework.
  In contrast, our \ourmodel~offers a collaborative way to leverage the appearance and motion cues under the \textbf{mutual restraint} of full-duplex strategy, thus providing more accurate structure details and alleviating the short-term feature drifting issue~\cite{yang2019anchor}.
  }\label{fig:front_figure_1}
\end{figure}

Intuitively, appearance and motion characteristics should be homogeneous to a certain degree for the same object within a short time. As seen in \figref{fig:front_figure_1}, the foreground region of appearance and motion intrinsically share the correlative patterns about perceptions, including semantic structure, movement trends.
Nevertheless, misguided knowledge in the individual modality, \eg, static shadow under the chassis and small car in the background, will produce inaccuracies during the feature propagation. Thus, it easily stains the result (see blue boxes).

To address these challenges, we introduce a novel modality transmission strategy (\textit{full-duplex}~\cite{bharadia2013full}) between spatial- and temporal-aware, instead of embedding them individually.
The proposed strategy is the bidirectional attention scheme across motion and appearance cues, which explicitly incorporates the appearance and motion patterns in a unified framework as depicted in~\figref{fig:teaser_figure}~(c).
As seen in~\figref{fig:front_figure_1}, our method visually performs better than the one with a simplex strategy (a \& b).

To fully investigate the simplex and full-duplex strategies of our framework, we present the following contributions: 
\begin{itemize}
\item We propose a unified framework Full-duplex Strategy Network (\textbf{\ourmodel}) for robust video object segmentation, which makes full use of spatial-temporal representations.

\item We adopt a bidirectional interaction module, dubbed the relational cross-attention module (RCAM), to extract discriminative features from the appearance and motion branches, which ensures mutual restraint between each other.
To improve the model robustness, we introduce a 
bidirectional purification module (BPM), which is equipped with an interlaced decremental connection to update inconsistent features between the spatial-temporal embeddings automatically.

\item We demonstrate that our \ourmodel~achieves favourable performance on five mainstream benchmarks, especially for our \ourmodel~($N$=4, CRF) outperforms the SOTA U-VOS model (\ie, MAT~\cite{zhou2020motion_attentive}) on the DAVIS$_{16}$~\cite{perazzi2016benchmark} leaderboard by a margin of 2.4\% in terms of Mean-$\mathcal{F}$ score, with less training data (\ie, Ours-13K \textit{vs.} MAT-16K). 
\end{itemize}

As an extension of our ICCV-2021 version~\cite{ji2021FSNet}, we incorporate more details to provide a better understanding of our novel framework as follow:   
\begin{itemize}
    \item To provide our community with a comprehensive study, we have made a lot of efforts to improve the presentations (\eg, \figref{fig:teaser_figure}, \figref{fig:front_figure_1}, and \figref{Fig:performance1}) and discussions (see \secref{sec:discussion}) of our manuscript.
    
    \item We investigate the self-purification mode of BPM under our \ourmodel~(see~\figref{fig:self_purf} and~\secref{sec:self_purf}), the relation between RCAM and BPM (see~\secref{sec:relation_RCAM_BPM}), and the training effectiveness with less data (see~\secref{sec:scale_of_dataset}). The results further demonstrate the validity and rationality of our current design under various conditions.
    
    \item We provide more details to the conference version, such as backbone details (see~\secref{sec:backbone_details}), evaluation metrics (see~\secref{sec:metric}), prediction selection (see~\secref{sec:prediction_selection}), and post-processing techniques (see~\secref{sec:crf_post_process}).
    
    \item We further present more performances under different thresholds used in the final results (\ie, PR curve in \figref{fig:pr_curve}). Additional test results (\ie, DAVSOD$_{19}$-Normal25, DAVSOD$_{19}$-Difficult20) on a recent challenging dataset still show that our framework is superior to existing SOTA models (see~\tabref{tab:davsod_benchmark}). 
\end{itemize}


\section{Related Works}
Depending on whether or not the first frame of ground truth is given, the VOS task can be divided into two scenarios, \ie, \textit{semi-supervised}~\cite{wang2018semi} and \textit{unsupervised/zero-shot}~\cite{bmvc2020making}.
Some typical semi-supervised VOS models can be referred to~\cite{zhang2020transductive,robinson2020learning,zhang2020dual,ijcai2020E3SN,li2020fast,seong2020kernelized,bhat2020learning,yang2019video,xu2018youtube,caelles20192019,hu2021learning,duke2021sstvos,johnander2019generative,wug2018fast,voigtlaender2019feelvos,cheng2018fast,caelles2017one,perazzi2017learning}.
This paper studies the unsupervised setting~\cite{zhou2020motion_attentive,zhou2020matnet}, leaving the semi-supervised setting as our future work. 
\subsection{Unsupervised VOS}
Although there are many works addressing the VOS task in a semi-supervised manner, \ie, by supposing an object mask annotation is given in the first frame, other researchers have attempted to address the more challenging unsupervised VOS (U-VOS) problem. 
Early U-VOS models resort to low-level handcrafted features for heuristic segmentation inference, such as long sparse point trajectories~\cite{brox2010object,ochs2012higher,fragkiadaki2012video,shi1998motion,wang2017super}, object proposals~\cite{lee2011key,li2013video,ma2012maximum,perazzi2015fully}, saliency priors~\cite{wang2015saliency,faktor2014video,wang2015robust}, optical flow~\cite{tsai2016video}, or superpixels~\cite{galasso2012video,grundmann2010efficient,xu2012streaming}.
These traditional models have limited generalizability and thus low accuracy in highly dynamic and complex scenarios due to their lack of semantic information and high-level content understanding. Recently, RNN-based models~\cite{song2018pyramid,wang2019learning,xu2018youtube,zheng2019cascaded,tokmakov2017learning1,ballas2015delving} have become popular due to their better capability of capturing long-term dependencies and their use of deep learning.
In this case, U-VOS is formulated as a recurrent modelling issue over time, where spatial features are jointly exploited with long-term temporal context.

\begin{figure*}[t!]
  \centering
  \includegraphics[width=.99\linewidth]{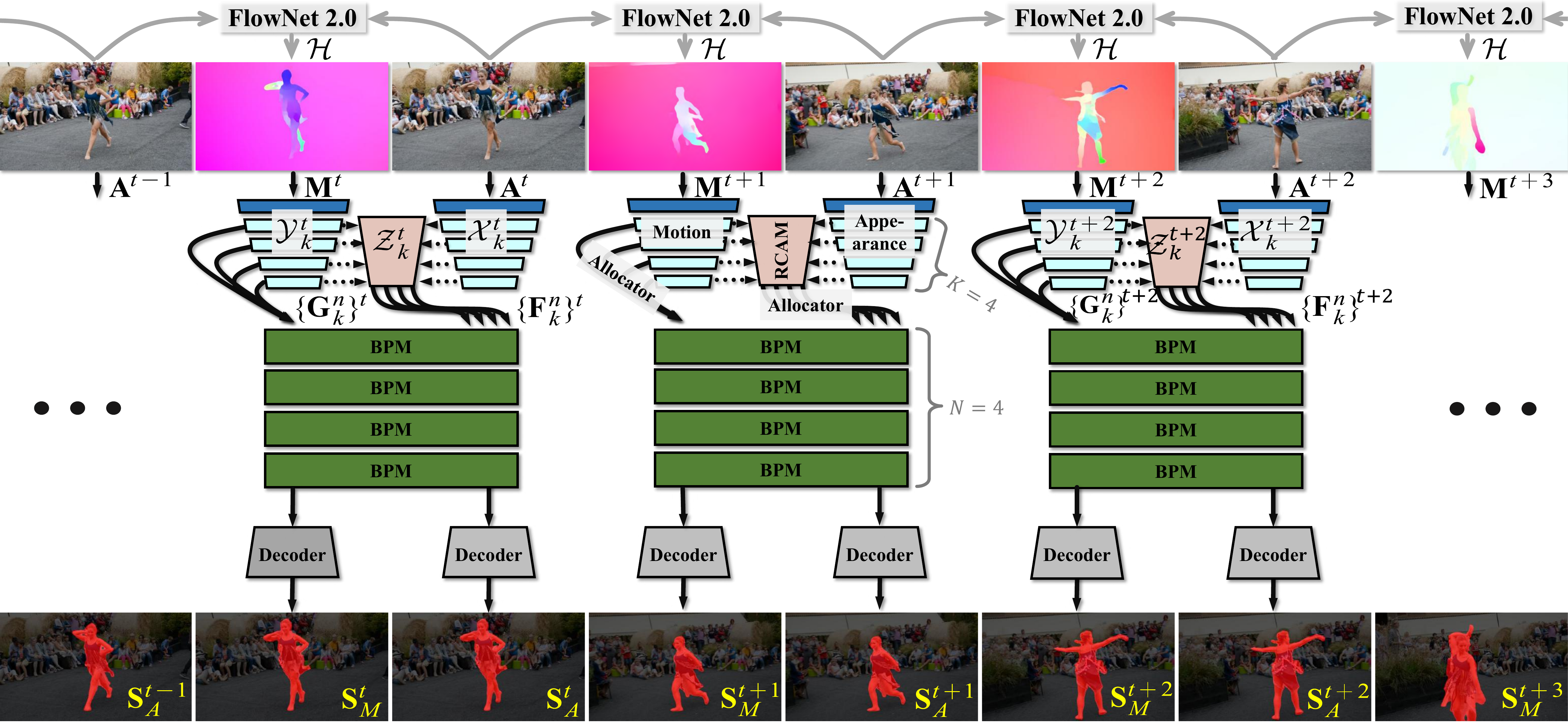}
  \caption{The architecture of our \ourmodel~for video object segmentation. The Relational Cross-Attention Module (RCAM) abstracts more discriminative representations between the motion and appearance cues using the full-duplex strategy.
  Then four Bidirectional Purification Modules (BPM) are stacked to further re-calibrate inconsistencies between the motion and appearance features. Finally, we utilize the decoder to generate our prediction.
  }
  \label{Fig:overall_framework}
\end{figure*}

How to combine motion cues with appearance features is a long-standing problem in this field. To this end, Tokmakov~\etal~\cite{tokmakov2017learning2} proposed to use the motion patterns required from the video simply. However, their method cannot accurately segment objects between two similar consecutive frames since it relies heavily on the guidance of optical flow.
To resolve this, several works~\cite{tokmakov2017learning1,cheng2017segflow,siam2019video} have integrated the spatial and temporal features from the parallel network, which can be viewed as plain feature fusion from the independent spatial and temporal branch with an implicit modelling strategy.
Li \etal~\cite{li2018unsupervisedECCV} proposed a multi-stage processing method to tackle U-VOS, which first utilizes a fixed appearance-based network to generate objectness and then feeds this into the motion-based bilateral estimator to segment the objects.

\subsection{Attention-based VOS}
The attention-based VOS task is closely related to U-VOS since it extracts attention-aware object(s) from a video clip.
Traditional methods~\cite{wang2017saliency,zhou2018improving,xu2019video,hu2018unsupervised} first compute the single-frame saliency based on various hand-crafted static and motion features and then conduct spatial-temporal optimization to preserve coherency across consecutive frames.
Recent works~\cite{wang2017videoTIP,le2017deeply,min2019tased} aim to learn a highly semantic representation and usually perform spatial-temporal detection end-to-end.
Many schemes have been proposed to employ deep networks that consider temporal information, such as ConvLSTM~\cite{song2018pyramid,fan2019shifting,li2018flow}, take optical-flows/adjacent-frames as input~\cite{li2019motion,wang2017videoTIP}, 3D convolutional~\cite{le2017deeply,min2019tased}, or directly exploit temporally concatenated deep features~\cite{le2018video}.
Besides, long-term influences are often taken into account and combined with deep learning. Li~\etal~\cite{li2019accurate} proposed a key-frame strategy to locate representative high-quality video frames of salient objects~\cite{borji2019salient,zhou2021rgb} and diffused their saliency to ill-detected non-key frames.
Chen~\etal~\cite{chen2019improved} improved saliency detection by leveraging long-term spatial-temporal information, where high-quality ``beyond-the-scope frames'' are aligned with the current frames. Both types of information are fed to deep neural networks for classification. 
Besides considering how to better leverage temporal information, other researchers have attempted to address different problems in video salient object detection (V-SOD), such as reducing the data labelling requirements~\cite{yan2019semi}, developing semi-supervised approaches~\cite{tang2018weakly}, or investigating relative saliency~\cite{wang2019ranking}. 
Fan~\etal~\cite{fan2019shifting} recently introduced a V-SOD model equipped with a saliency shift-aware ConvLSTM, together with an attention-consistent V-SOD dataset with high-quality annotations.
Zhao~\etal~\cite{zhao2021weakly} build a large-scale with scribble annotation for weakly supervised video salient object detection. They propose an appearance-motion fusion module to aggregate the spatial-temporal features attentively.

\section{Methodology}\label{Sec:methodology}

\subsection{Overview}\label{Sec:framework_overivew}
Suppose that a video clip contains $T$ consecutive frames $\{\mathbf{A}^t\}_{t=1}^{T}$.
We first utilize optical flow field generator $\mathcal{H}$, \ie, FlowNet 2.0~\cite{ilg2017flownet}, to generate $T-1$ optical flow maps $\{\mathbf{M}^{t}\}_{t=1}^{T-1}$, which are all computed by two adjacent frames ($\mathbf{M}^t = \mathcal{H}[\mathbf{A}^{t},\mathbf{A}^{t+1}]$).
To ensure the inputs match, we discard the last frame in the pipeline.
Thus, the proposed pipeline takes both the appearance image $\{\mathbf{A}^{t}\}_{t=1}^{T-1}$ and its paired motion map $\{\mathbf{M}^{t}\}_{t=1}^{T-1}$ as the input.
First, $\mathbf{M}^{t}$ \& $\mathbf{A}^{t}$ pairs at frame $t$\footnote{Here, we omit the superscript ``$t$'' for the convenient expression.} are fed to two independent ResNet-50~\cite{he2016deep} branches (\ie, motion and appearance blocks in \figref{Fig:overall_framework}). 
The appearance features $\{\mathcal{X}_k\}_{k=1}^{K}$ and motion features $\{\mathcal{Y}_k\}_{k=1}^{K}$ extracted from $K$ layers are then sent to the Relational Cross-Attention Modules (RCAMs), which allows the network to embed spatial-temporal cross-modal features.
%
Next, we employ the Bidirectional Purification Modules (BPMs) with $N$ cascaded units. BPMs focus on distilling representative carriers from fused features $\{\mathbf{F}_k^n\}^{N}_{n=1}$ and motion-based features $\{\mathbf{G}_k^n\}_{n=1}^{N}$.
Finally, the predictions (\ie, $\mathbf{S}_M^t$ and $\mathbf{S}_A^t$) at frame $t$ are generated from two decoder blocks.

\begin{figure*}[t!]
  \centering
  \includegraphics[width=.8\textwidth]{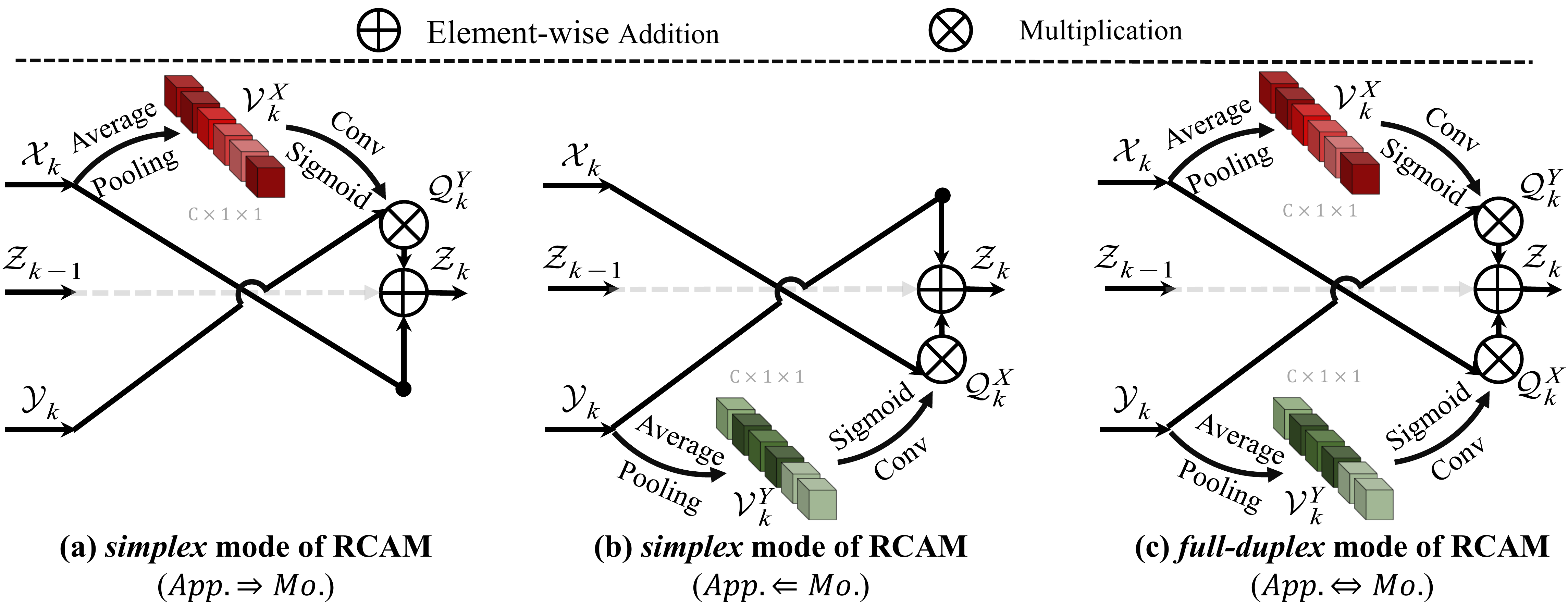}
  \caption{Illustration of our Relational Cross-Attention Module (RCAM) with a \textit{simplex} (a~\&~b) and \textit{full-duplex} (c) strategy. 
  }\label{Fig:framework_RCAM}
\end{figure*}

\subsection{Relational Cross-Attention Module}
As discussed in~\secref{sec:introduction}, a single-modality (\ie, motion or appearance) guided stimulation may cause the model to make incorrect decisions.
To alleviate this, we design a cross-attention module (RCAM) via the channel-wise attention mechanism, which focuses on distilling out effectively squeezed cues from two modalities and then modulating each other.
As shown in~\figref{Fig:framework_RCAM} (c), the two inputs of RCAM are appearance features $\{\mathcal{X}_k\}_{k=1}^{K}$ and motion features $\{\mathcal{Y}_k\}_{k=1}^{K}$, which are obtained from the two different branches of the standard ResNet-50~\cite{he2016deep}.
Specifically, for each \textit{k}-level, we first perform global average pooling (GAP) to generate channel-wise vectors $\mathcal{V}^{X}_{k}$ and $\mathcal{V}^{Y}_{k}$ from each $\mathcal{X}_k$ and $\mathcal{Y}_k$. 
Next, two 1$\times$1 conv, \ie, $\phi(x;\mathbf{W}_\phi)$ and $\theta(x;\mathbf{W}_\theta)$, with learnable parameters $\mathbf{W}_\phi$ and $\mathbf{W}_\theta$, generate two discriminated global descriptors.
The sigmoid function $\sigma[x]=e^{x}/(e^{x}+1),~x\in\mathbb{R}$ is then applied to convert the final descriptors into the interval [0, 1], \ie, into the valid attention vector for channel weighting.
Then, we perform outer product $\otimes$ between $\mathcal{X}_k$ and $\sigma\left[\theta(\mathcal{V}_{k}^{Y};\mathbf{W}_\theta)\right]$ to generate a candidate feature $\mathcal{Q}_{k}^{X}$, and vice versa, as follows:
\begin{equation}
     \mathcal{Q}_{k}^{X} = \mathcal{X}_k \otimes \sigma\left[\theta(\mathcal{V}_{k}^{Y};\mathbf{W}_\theta)\right],
\end{equation}
\begin{equation}
     \mathcal{Q}_{k}^{Y} = \mathcal{Y}_k \otimes \sigma\left[\phi(\mathcal{V}_{k}^{X};\mathbf{W}_\phi)\right].
\end{equation}

Then, we combine $\mathcal{Q}_{k}^{X}$, $\mathcal{Q}_{k}^{Y}$, and lower-level fused feature $\mathcal{Z}_{k-1}$ for in-depth feature extraction.
With an element-wise addition operation $\oplus$, conducted in the corresponding \textit{k}-th level block $\mathcal{B}_k[x]$ in the ResNet-50, we finally obtain the fused features $\mathcal{Z}_k$ that contain comprehensive spatial-temporal correlations:
\begin{equation}\label{equ:3}
  \mathcal{Z}_k = \mathcal{B}_k \left[ \mathcal{Q}^{X}_{k} \oplus \mathcal{Q}^{Y}_{k} \oplus \mathcal{Z}_{k-1} \right],
\end{equation}
where $k \in \{1:K\}$ denotes different feature hierarchies in the backbone.
Note that $\mathcal{Z}_0$ denotes the zero tensor.
In our implementation, we use the top four feature pyramids, \ie, $K=4$, suggested by~\cite{wei2019f3net,zhang2018exfuse}.

\subsection{Bidirectional Purification Module}
In addition to the RCAM described above, which integrates common cross-modality features, we further introduce the bidirectional purification module (BPM) to improve the model robustness. 
Following the standard in action recognition~\cite{sevilla2018integration} and saliency detection~\cite{wu2019stacked}, our bidirectional purification phase comprises $N$ BPMs connected in a cascaded manner. 
As shown in Fig.~\ref{Fig:overall_framework}, we first employ the feature allocator $\psi_{\{F,G\}}(x;\mathbf{W}_\psi^{\{F,G\}})$ to unify the feature representations from the previous stage:
\begin{equation}\label{equ:allocator}
    \mathbf{F}_{k}^{n} = \psi_{F}(\mathcal{Z}_k;\mathbf{W}_\psi^{F}),~
    \mathbf{G}_{k}^{n} = \psi_{G}(\mathcal{Y}_k;\mathbf{W}_\psi^{G}),
\end{equation}
where $k \in \{1:K\}$ and $n \in \{1:N\}$ denote different feature hierarchies and number of BPM, respectively.
To be specific, $\psi_{\{F,G\}}(x;\mathbf{W}_{\psi}^{\{F,G\}})$ is composed of two 3$\times$3 conv, each with 32 filters to reduce the feature channels. 
Note that the allocator is conducive to reduce the computational burden as well as facilitate various element-wise operations. 

\begin{figure*}[t!]
  \centering
  \includegraphics[width=.9\textwidth]{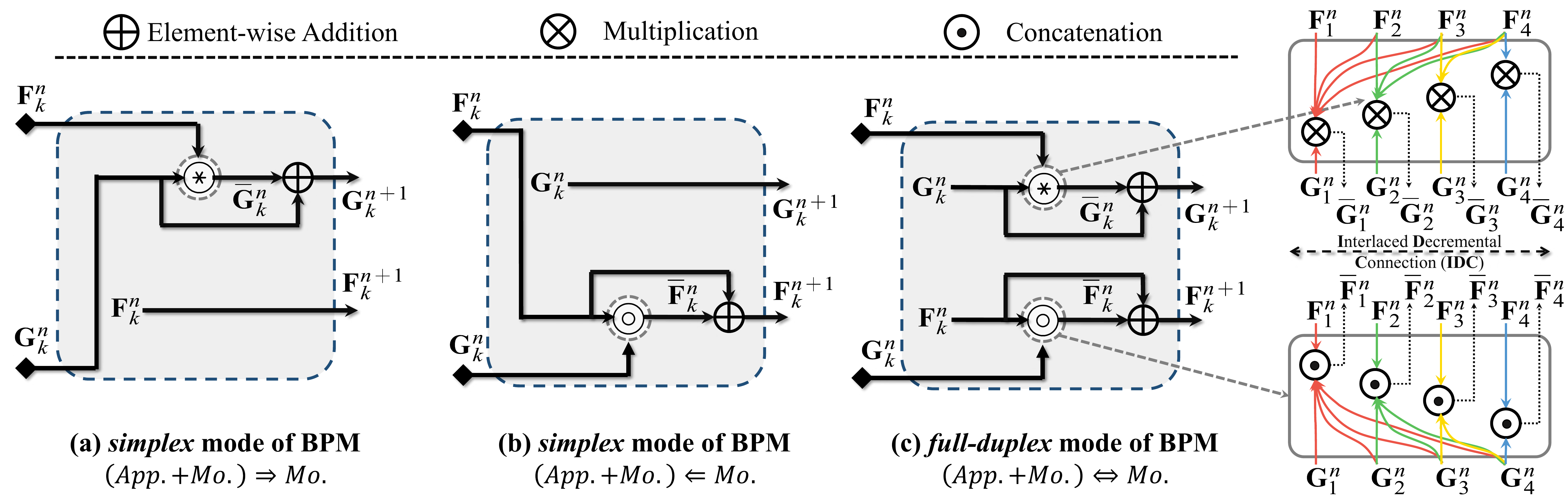}
  \caption{Illustration of our Bidirectional Purification Module (BPM) with a \textit{simplex} and \textit{full-duplex} strategy.
  }\label{Fig:framework_bpm}
\end{figure*}

%
%
Here, we consider a \textit{bidirectional attention} scheme (see \figref{Fig:framework_bpm} (c)) that contains two \textit{simplex} strategies (see \figref{Fig:framework_bpm} (a \& b)) in the BPM.
On the one hand, the motion features $\mathbf{G}_{k}^{n}$ contain temporal cues and can be used to enrich the fused features $\mathbf{F}_{k}^{n}$ by the concatenation operation. 
On the other, the distractors in the motion feature $\mathbf{G}_{k}^{n}$ can be suppressed by multiplicating the fused features $\mathbf{F}_{k}^{n}$.
%
Besides acquiring robust feature representation, we introduce an efficient cross-modal fusion strategy in this scheme, which broadcasts high-level, semantically strong features to low-level, semantically weak features via interlaced decremental connection (IDC) with a top-down pathway~\cite{lin2017feature}.
Specifically, as the first part, the spatial-temporal feature combination branch (see \figref{Fig:framework_bpm} (b)) is formulated as:
\begin{equation}\label{equ:5}
\mathbf{F}_{k}^{n+1} = \mathbf{F}_{k}^{n} \oplus \bigcup_{i=k}^{K} \left[ \mathbf{F}_{k}^{n},\mathcal{P} ( \mathbf{G}_{i}^{n}) \right],
\end{equation}
where $\mathcal{P}$ is an up-sampling operation followed by a 1$\times$1 convolutional layer (conv) to reshape the candidate guidance to a consistent size with $\mathbf{F}_{k}^{n}$. 
Symbols $\oplus$ and $\bigcup$ respectively denote element-wise addition and concatenation operations with an IDC strategy\footnote{For instance, $\mathbf{\bar{G}}_{2}^{n} = \bigcup_{i=2}^{K=4} [\mathbf{F}_{2}^{n}, \mathcal{P} ( \mathbf{G}_{i}^{n})] = \mathbf{F}_{2}^{n} \odot \mathcal{P} ( \mathbf{G}_{2}^{n}) \odot \mathcal{P} (\mathbf{G}_{3}^{n}) \odot \mathcal{P} ( \mathbf{G}_{4}^{n})$ when $k=2$ and $K=4$.}, followed by a 1$\times$1 conv with 32 filters.
For the other part, we formulate the temporal feature re-calibration branch (see \figref{Fig:framework_bpm} (a)) as:
\begin{equation}
\mathbf{G}_{k}^{n+1} = \mathbf{G}_{k}^{n} \oplus \bigcap_{j=k}^{K}[ \mathbf{G}_{k}^{n},\mathcal{P}(\mathbf{F}_{j}^{n})],
\end{equation}
where $\bigcap$ denotes element-wise multiplication with an IDC strategy, followed by a 1$\times$1 conv with 32 filters.
%

\subsection{Decoder}\label{sec:decoder}
After feature aggregation and re-calibration with multi-pyramidal interaction, the last BPM unit produces two groups of discriminative features (\ie, $\mathbf{F}_{k}^{N}$ \& $\mathbf{G}_{k}^{N}$) with a consistent channel number of 32.
We integrate pyramid pooling module (PPM)~\cite{zhao2017pyramid} into each skip connection of the U-Net~\cite{ronneberger2015u} as our decoder, and only adopt the top four layers in our implementation ($K=4$).
Since the features are fused from high to low level, global information is well retained at different scales of the designed decoder:
\begin{equation}\label{equ:ppm_1}
    \hat{\mathbf{F}}_{k}^{N} = \mathcal{C} [ \mathbf{F}_{k}^{N} \odot \mathcal{UP}(\hat{\mathbf{F}}_{k+1}^{N}) ],
\end{equation}
\begin{equation}\label{equ:ppm_2}
    \hat{\mathbf{G}}_{k}^{N} = \mathcal{C} [ \mathbf{G}_{k}^{N} \odot \mathcal{UP}(\hat{\mathbf{G}}_{k+1}^{N}) ].
\end{equation}
Here, $\mathcal{UP}$ indicates the upsampling operation after the pyramid pooling layer, while $\odot$ is a concatenation operation between two features. Then, a conv $\mathcal{C}$ is used for reducing the channels from 64 to 32. 
Lastly, we use a 1$\times$1 conv with a single filter after the upstream output (\ie, $\hat{\mathbf{F}}_{1}^{N}$ \& $\hat{\mathbf{G}}_{1}^{N}$), followed by a sigmoid activation function to generate the predictions (\ie, $\mathbf{S}_{A}^{t}$ \& $\mathbf{S}_{M}^{t}$) at frame $t$.

\subsection{Learning Objective}\label{sec:training}
Given a group of predictions $\mathbf{S}^{t} \in \{\mathbf{S}_{A}^{t}, \mathbf{S}_{M}^{t}\}$ and the corresponding ground-truths $\mathbf{G}^{t}$ at frame $t$, we employ the standard binary \textit{cross-entropy} loss $\mathcal{L}_{bce}$ to measure the dissimilarity between output and target, which computes:
\begin{equation}
\begin{aligned}
    \mathcal{L}_{bce}(\mathbf{S}^{t},\mathbf{G}^{t}) = 
    &-\sum_{(x, y)}[\mathbf{G}^{t}(x, y) \log (\mathbf{S}^{t}(x, y)) \\ 
    &+ (1 - \mathbf{G}^{t}(x, y)) \log (1-\mathbf{S}^{t}(x, y))],
\end{aligned}
\end{equation}
where $(x,y)$ indicates a coordinate in the frame. The overall loss function is then formulated as:
\begin{equation}
    \mathcal{L}_{total} = \mathcal{L}_{bce}(\mathbf{S}_{A}^{t},\mathbf{G}^{t}) + \mathcal{L}_{bce}(\mathbf{S}_{M}^{t},\mathbf{G}^{t}).
\end{equation}
For final prediction, we use $\mathbf{S}_{A}^{t}$ since our experiments show that it performs better when combining appearance and motion cues. 

\subsection{Implementation Details}~\label{sec:implementation_details}

\subsubsection{Backbone Details}\label{sec:backbone_details}
Without any modification, three standard ResNet-50~\cite{he2016deep} (removing the top-three layers: average pooling, fully-connected, and softmax layers) backbones are adopted for the appearance branch, the motion branch and the merging branch.
Each ResNet-50 backbone results in $K=4$ hierarchies inspired by previous work~\cite{wei2019f3net}. 
After removing the top fully connected layers, the feature hierarchies ($\{\mathcal{X}_k, \mathcal{Y}_k, \mathcal{Z}_k\}, k \in \{2:5\}$) from shallow to deep are extracted from the conv2\_3 ($k=2$), conv3\_4 ($k=3$), conv4\_6 ($k=4$), and conv5\_3 ($k=5$) layers of the ResNet-50, respectively. 
Note that we have also tried a two-branches setting, namely removing the merging branch and letting $\mathcal{Z}_k = \mathcal{Q}^X_k \oplus \mathcal{Q}^{Y}_{k} \oplus \mathcal{Z}_{k-1}$ instead of $\mathcal{Z}_k = \mathcal{B}_k [ \mathcal{Q}_k^X \oplus \mathcal{Q}^Y_x \oplus \mathcal{Z}_{k-1}]$ in~\equref{equ:3}.
Unfortunately, this leads to a 2.5\% performance drop in performance concerning $S_\alpha$ on the DAVIS$_{16}$~\cite{perazzi2016benchmark} dataset.
This is because the third merging branch can sequentially enhance and promote the spatial-temporal features from RCAMs, leading to better segmentation accuracy.

\subsubsection{Training Settings}\label{sec:training_scheme}
We implement our model in PyTorch~\cite{paszke2019pytorch}, accelerated by an NVIDIA RTX TITAN GPU. 
All the inputs are uniformly resized to 352$\times$352. 
To enhance the stability and generalizability of our learning algorithm, we employ the multi-scale (\ie, $\{0.75,1,1.25\}$) training strategy~\cite{he2015spatial} in the training phase. 
As can be seen from the experimental results in~\tabref{tab:number_of_bpm}, the variant with $N$=4 (the number of BPM) achieves the best performance.
We utilize the stochastic gradient descent (SGD) algorithm to optimize the entire network, with a momentum of $0.9$, a learning rate of $2e^{-3}$, and a weight decay of $5e^{-4}$.
The learning rate decreased by 10\% per 20 epochs.

\subsubsection{Testing Settings and Runtime}
Given a frame along with its motion map, we resize them to 352$\times$352 and feed them into the corresponding branch.
Similar to~\cite{zhou2020motion_attentive,lu2019see,wang2019learning}, We employ the conditional random field (CRF)~\cite{krahenbuhl2011efficient} post-processing technique.
%
%
The inference time of our method is 0.08s per frame, regardless of flow generation and CRF post-processing.

\section{Experiments}

\subsection{Experimental Protocols}

\subsubsection{Datasets}
We evaluate the proposed model on four widely used VOS datasets. 
DAVIS$_{16}$~\cite{perazzi2016benchmark} is the most popular of these, and consists of 50 (30 training and 20 validation) high-quality and densely annotated video sequences.
MCL~\cite{kim2015spatiotemporal} contains 9 videos and is mainly used as testing data.
FBMS~\cite{ochs2013segmentation} includes 59 natural videos, in which 29 sequences are 
used as the training set and 30 are for testing. 
SegTrack-V2~\cite{li2013video} is one of the earliest VOS datasets and consists of 13 clips. 
In addition, DAVSOD$_{19}$~\cite{fan2019shifting} was specifically designed for the V-SOD task. 
It is the most challenging visual attention consistent V-SOD dataset with high-quality annotations and diverse attributes. 

\subsubsection{Training}
Following a similar multi-task training setup as~\cite{li2019motion}, we divide our training procedure into three steps:
\begin{itemize}
    \item We first adopt a well-known static saliency dataset DUTS~\cite{wang2017learning} to train the spatial branch to avoid over-fitting, like in~\cite{wang2017videoTIP,song2018pyramid,fan2019shifting}. This step lasts for 50 epoches with a batch size of 8 under the same training settings mentioned in~\secref{sec:training_scheme}.
    \item We then train the temporal branch on the generated optical flow maps. This step lasts for 50 epoches with a batch size of 8 under the same training settings mentioned in~\secref{sec:training_scheme}.
    \item We finally load the weights pre-trained on two sub-tasks into the spatial and temporal branches, and thus, the whole network is end-to-end trained on the training set of DAVIS$_{16}$ (30 clips) and FBMS (29 clips). The last step takes about 4 hours and converges after 20 epochs with a batch size of 8 under same the training settings mentioned in~\secref{sec:training_scheme}.
\end{itemize}

\subsubsection{Testing} 
We follow the standard benchmarks~\cite{perazzi2016benchmark,fan2019shifting} to test our model on the validation set (20 clips) of DAVIS$_{16}$, the test set of FBMS (30 clips), the test set (Easy35 split) of DAVSOD$_{19}$ (35 clips), the whole of MCL (9 clips), and the whole of SegTrack-V2  (13 clips).

\subsection{Evaluation Metrics}\label{sec:metric}

We define a prediction map at frame $t$ as $\mathbf{S}_{A}^{t}$ and its corresponding ground-truth mask as $\mathbf{G}^t$. The formulations of the metrics are given as follows.

\subsubsection{Metrics for U-VOS task}
Following~\cite{yang2019anchor}, we utilize two standard metrics to evaluate the performance of U-VOS models. Note that all prediction maps are ensured to be binary in the U-VOS task.

\begin{enumerate}
    \item \textbf{Mean Region Similarity}: This metric, also called jaccard similarity coefficient, is defined as the intersection-over-union of the prediction map and the ground-truth mask. The formulation is defined as:
    \begin{equation}\label{equ:region_sim}
      \mathcal{J} = \frac{|\mathbf{S}_{A}^{t} \cap \mathbf{G}^t|}{|\mathbf{S}_{A}^{t} \cup \mathbf{G}^t|},
    \end{equation}
    where $|\cdot|$ is the number of pixels in the area. In all of our experiments, we also report the mean value of Mean-$\mathcal{J}$, similar to~\cite{yang2019anchor}.
    
    \item \textbf{Mean Contour Accuracy}: Here, the contour accuracy metric we used is also called the contour F-measure. We compute the contour-based precision and recall between the contour points of $c(\mathbf{S}_{A}^{t})$ and $c(\mathbf{G}^t)$, where $c(\cdot)$ is the extraction of contour points of a mask. The formulation is defined as:
    \begin{equation}\label{equ:contour}
      \mathcal{F} = \frac{2 \times \textrm{Precision}_{c} \times \textrm{Recall}_{c}}{\textrm{Precision}_c + \textrm{Recall}_c},
    \end{equation}
    where $\textrm{Precision}_{c} = |c(\mathbf{S}_{A}^{t}) \cap c(\mathbf{G}^t)|/|c(\mathbf{S}_{A}^{t})|$, $\textrm{Recall}_c = |c(\mathbf{S}_{A}^{t}) \cap c(\mathbf{G}^t))| / |c(\mathbf{G}^t)|$.
    Similar to~\cite{yang2019anchor}, we also report the mean value of Mean-$\mathcal{F}$ in all of our experiments.
\end{enumerate}

\subsubsection{Metrics for V-SOD task}\label{sec:metric_vsod}
Unlike the U-VOS task, the prediction map can be non-binary in the V-SOD benchmarking. More details refer to \secref{sec:evaluation_on_davis16}.

\begin{enumerate}
    \item \textbf{Mean Absolute Error (MAE)}: It is a typical pixel-wise measure, which is defined as:
    \begin{equation}\label{equ:mae}
      \mathcal{M} = \frac{1}{W \times H} \sum_{x}^{W} \sum_{y}^{H} |\mathbf{S}_{A}^{t}(x,y) - \mathbf{G}^t(x,y)|,
    \end{equation}
    where $W$ and $H$ are the width and height of ground-truth $\mathbf{G}^t$, and $(x,y)$ are the coordinates of a pixel in $\mathbf{G}^t$.
    
    \item \textbf{Precision-Recall (PR) Curve}: Precision and recall~\cite{Achanta2009,cheng2015global,Borji2015TIP} are defined as:
    \begin{equation}
        \textrm{Precision}=\frac{|\mathbf{S}_{A}^{t}(T) \cap \mathbf{G}^t|}{|\mathbf{S}_{A}^{t}(T)|},
    \end{equation}
    \begin{equation}
        \textrm{Recall} =\frac{|\mathbf{S}_{A}^{t}(T)\cap \mathbf{G}^t|}{|\mathbf{G}^t|},
    \end{equation}
    where $\mathbf{S}_{A}^{t}(T)$ is the binary mask obtained by directly thresholding the prediction map $\mathbf{S}_{A}^{t}$ with the threshold $T \in [0,255]$, and $|\cdot|$ is the total area of the mask inside the map. By varying $T$, a precision-recall curve can be obtained.
    
    \item \textbf{Maximum F-measure}: This is defined as:
    \begin{equation}\label{equ:max_f}
      F_\beta = \frac{(1+\beta^2) \textrm{Precision} \times \textrm{Recall}} {\beta^{2} \times \textrm{Precision} + \textrm{Recall}},
    \end{equation}
    where $\beta^2$ is set to 0.3 to focus more on the precision value than the recall value, as recommended in~\cite{borji2015salient}. We convert the non-binary prediction map into binary masks with threshold values from 0 to 255. In this paper, we report the maximum (\ie, $F_{\beta}^{max}$)  of a series of F-measure values calculated from the precision-recall curve by iterating over all the thresholds.
    
    \item \textbf{Maximum Enhanced-Alignment Measure}: As a recently proposed metric, $E_{\xi}$~\cite{achanta2009frequency} is used to evaluate both the local and global similarity between two binary maps. The formulation is as follows:
    \begin{equation}\label{equ:e_m}
      E_{\xi} = \frac{1}{W \times H} \sum_{x}^{W} \sum_{y}^{H} \phi\left[\mathbf{S}_{A}^{t}(x, y), \mathbf{G}^t(x, y)\right],
    \end{equation}
    where $\phi$ is the enhanced-alignment matrix. Similar to $F_\beta^{max}$, we report the maximum $E_{\xi}$ value computed from all the thresholds in all of our comparisons.
    
    \item \textbf{Structure Measure}: Fan~\etal~\cite{fan2017structure} proposed a metric to measure the structural similarity between a non-binary saliency map and a ground-truth mask:
    \begin{equation}\label{equ:s_m}
      S_\alpha = (1-\alpha) \times S_o(\mathbf{S}_{A}^{t}, \mathbf{G}^t) + \alpha \times S_r(\mathbf{S}_{A}^{t}, \mathbf{G}^t),
    \end{equation}
    where $\alpha$ balances the object-aware similarity $S_o$ and region-aware similarity $S_r$. We use the default setting ($\alpha=0.5$) suggested in \cite{fan2017structure}.

\end{enumerate}

\begin{table*}[t!]
\centering
  \footnotesize
  \renewcommand{\arraystretch}{1.3}
  \setlength\tabcolsep{0.9pt}
  \caption{Video object segmentation (VOS) performance of our~\ourmodel, compared with 14 SOTA unsupervised models and seven semi-supervised models on DAVIS$_{16}$~\cite{perazzi2016benchmark} validation set. 
  `\textit{w/} Flow': the optical flow algorithm is used.
  `\textit{w/} CRF': conditional random field~\cite{krahenbuhl2011efficient} is used for post-processing.
  The best scores are marked in \textbf{bold}.}
  \label{tab:davis16}
  \begin{tabular}{cl|cc|cccccccccccccc|ccccccccccc}
  \toprule
   & & \multicolumn{16}{c|}{Unsupervised} & \multicolumn{7}{c}{Semi-supervised}\\
   \cline{3-25}
   \multirow{0}{*}{} & \multirow{0}{*}{Metrics}
       & \multicolumn{2}{c|}{\textbf{\ourmodel}}
       & MAT    & AGNN      & AnDiff    & COS    & AGS 
       & EpO+   & MOA       & LSMO      & ARP       & LVO 
       & LMP    & SFL       & ELM       & FST
       & CFBI 
       & AGA  & RGM   & FEEL  & FA	 & OS & MSK \\  
       & & \multicolumn{2}{c|}{\textbf{(Ours)}}
       & \cite{zhou2020motion_attentive}
       & \cite{Wang_2019_ICCV}
       & \cite{yang2019anchor} & \cite{lu2019see} & \cite{wang2019learning} & \cite{faisal2019exploiting} & \cite{siam2019video}
       & \cite{tokmakov2019learning}     & \cite{koh2017primary} & \cite{tokmakov2017learning1}
       & \cite{tokmakov2017learning2}    & \cite{cheng2017segflow}  & \cite{lao2018extending} & \cite{papazoglou2013fast}
       & \cite{yang2020collaborative} 
       & \cite{johnander2019generative}  & \cite{wug2018fast}   & \cite{voigtlaender2019feelvos}  & \cite{cheng2018fast}	 & \cite{caelles2017one}
       & \cite{perazzi2017learning} \\
  \hline
        & \textit{w/} Flow        
    & \checkmark  & \checkmark
    & \checkmark  &                 
    &                 &                 &                 & \checkmark  & \checkmark
    & \checkmark  & \checkmark  
    & \checkmark  & \checkmark  & \checkmark  & \checkmark  & \checkmark    & 
    &                 &                 &                 &                 &                   & \checkmark \\
        & \textit{w/} CRF         
    & \checkmark  &                 
    & \checkmark  & \checkmark  &                 & \checkmark
    & \checkmark  & \checkmark  & \checkmark  & \checkmark  & 
    & \checkmark  & \checkmark  & 
    &                 &                 &                &               & \checkmark
    &                 & \checkmark  &                 & \checkmark  \\
  \hline
        & Mean-$\mathcal{J} \uparrow$
    & \cellcolor{mygray}{\textbf{83.4}} & 82.1
    & 82.4 & 80.7
    & 81.7 & 80.5 & 79.7 & 80.6 & 77.2
    & 78.2 & 76.2 & 75.9
    & 70.0 & 67.4 & 61.8 & 55.8
    & \cellcolor{mygray}{\textbf{85.3}} & 81.5 & 81.5 & 81.1 & 82.4 & 79.8 & 79.7 \\
        & Mean-$\mathcal{F} \uparrow$
    & 83.1 & \cellcolor{mygray}{\textbf{83.3}}
    & 80.7 & 79.1
    & 80.5 & 79.5 & 77.4 & 75.5 & 77.4
    & 75.9 & 70.6 & 72.1
    & 65.9 & 66.7 & 61.2 & 51.1
    & \cellcolor{mygray}{\textbf{86.9}} & 82.2 & 82.0 & 82.2 & 79.5 & 80.6 & 75.4 \\
  \bottomrule
  \end{tabular}
\end{table*}

%
\begin{table*}[t!]
\centering
  \footnotesize
  \renewcommand{\arraystretch}{1.0}
  \setlength\tabcolsep{6pt}
  \caption{Video salient object detection (V-SOD) performance of our \ourmodel, compared with 13 SOTA models on three popular V-SOD datasets, including DAVIS$_{16}$~\cite{perazzi2016benchmark}, MCL~\cite{kim2015spatiotemporal}, and FBMS~\cite{ochs2013segmentation}.
  `$\dag$' denotes that we generate non-binary saliency maps without CRF~\cite{krahenbuhl2011efficient} for a fair comparison.
  `N/A' means the results are not available.}
  \label{tab:score_VOS}
  \begin{tabular}{rr|cccc|cccc|cccc}
    \toprule
   &  &\multicolumn{4}{c|}{DAVIS$_{16}$~\cite{perazzi2016benchmark}}&\multicolumn{4}{c|}{MCL~\cite{kim2015spatiotemporal}} 
  & \multicolumn{4}{c}{FBMS~\cite{ochs2013segmentation}}
   \\
   \cline{3-14}
   \multirow{-2}{*}{} & \multirow{1}{*}{Model}
       & $S_{\alpha}\uparrow$ & $E_{\xi}^{max}\uparrow$
       & $F_{\beta}^{max}\uparrow$ & $\mathcal{M}\downarrow$
       & $S_{\alpha}\uparrow$ & $E_{\xi}^{max}\uparrow$
       & $F_{\beta}^{max}\uparrow$ & $\mathcal{M}\downarrow$
       & $S_{\alpha}\uparrow$ & $E_{\xi}^{max}\uparrow$ 
       & $F_{\beta}^{max}\uparrow$ & $\mathcal{M}\downarrow$ 
       \\
  \hline
  \multirow{5}{*}{\begin{sideways}2018\end{sideways}}
  & MBN~\cite{li2018unsupervisedECCV}  
    & 0.887   & 0.966   & 0.862   & 0.031
    & 0.755   & 0.858   & 0.698   & 0.119
    & 0.857   & 0.892   & 0.816   & 0.047
    \\
  & FGRN~\cite{li2018flow} 
    & 0.838   & 0.917   & 0.783   & 0.043
    & 0.709   & 0.817   & 0.625   & 0.044
    & 0.809   & 0.863   & 0.767   & 0.088
    \\
  & SCNN~\cite{tang2018weakly}
    & 0.761   & 0.843   & 0.679   & 0.077
    & 0.730   & 0.828   & 0.628   & 0.054
    & 0.794   & 0.865   & 0.762   & 0.095
    \\
  & DLVS~\cite{wang2017videoTIP}
    & 0.802   & 0.895   & 0.721   & 0.055
    & 0.682   & 0.810   & 0.551   & 0.060
    & 0.794   & 0.861   & 0.759   & 0.091
    \\
  & SCOM~\cite{chen2018scom}
    & 0.814   & 0.874   & 0.746   & 0.055
    & 0.569   & 0.704   & 0.422   & 0.204
    & 0.794   & 0.873   & 0.797   & 0.079
    \\
  \hline
  \multirow{9}{*}{\begin{sideways}2019$\sim$2020\end{sideways}}
  & RSE~\cite{xu2019video}
    & 0.748   & 0.878   & 0.698   & 0.063
    & 0.682   & 0.657   & 0.576   & 0.073
    & 0.670   & 0.790   & 0.652   & 0.128
    \\
  & SRP~\cite{cong2019video}
    & 0.662   & 0.843   & 0.660   & 0.070
    & 0.689   & 0.812   & 0.646   & 0.058
    & 0.648   & 0.773   & 0.671   & 0.134
    \\
  & MESO~\cite{xu2019TMMvideo}
    & 0.718   & 0.853   & 0.660   & 0.070
    & 0.477   & 0.730   & 0.144   & 0.102
    & 0.635   & 0.767   & 0.618   & 0.134
    \\
  & LTSI~\cite{chen2019improved}
    & 0.876   & 0.957   & 0.850   & 0.034
    & 0.768   & 0.872   & 0.667   & 0.044
    & 0.805   & 0.871   & 0.799   & 0.087
    \\
  & SPD~\cite{li2019accurate}
    & 0.783   & 0.892   & 0.763   & 0.061
    & 0.685   & 0.794   & 0.601   & 0.069
    & 0.691   & 0.804   & 0.686   & 0.125
    \\
  & SSAV\cite{fan2019shifting}
    & 0.893   & 0.948   & 0.861   & 0.028
    & 0.819   & 0.889   & 0.773   & 0.026
    & 0.879   & 0.926   & 0.865   & 0.040
    \\	
  & RCR\cite{yan2019semi}
    & 0.886   & 0.947   & 0.848   & 0.027
    & 0.820   & 0.895   & 0.742   & 0.028
    & 0.872   & 0.905   & 0.859   & 0.053
    \\
  & PCSA\cite{gu2020pyramid} 
    & 0.902   & 0.961   & 0.880   & 0.022
    & N/A     &  N/A    & N/A     & N/A
    & 0.868   & 0.920   & 0.837   & \textbf{0.040}
    \\
    \hline
  \rowcolor{mygray}
  &\textbf{\ourmodel$^\dag$} 
    & \textbf{0.920}   & \textbf{0.970}   & \textbf{0.907}   & \textbf{0.020}
    & \textbf{0.864}   & \textbf{0.924}   & \textbf{0.821}   & \textbf{0.023}
    & \textbf{0.890}   & \textbf{0.935}   & \textbf{0.888}   & 0.041
    \\	
  \bottomrule
  \end{tabular}
\end{table*}

\subsection{U-VOS and V-SOD tasks}

\subsubsection{Evaluation on DAVIS$_{16}$ dataset}\label{sec:evaluation_on_davis16}
As shown in~\tabref{tab:davis16}, we compare our \ourmodel~with 14 SOTA U-VOS models on the DAVIS$_{16}$ public leaderboard.
%
%
We also compare it with seven recent semi-supervised approaches as reference. 
We use a threshold of 0.5 to generate the final binary maps for a fair comparison, as recommended by~\cite{yang2019anchor}. 
Our \ourmodel~outperforms the best model (AAAI'20-MAT~\cite{zhou2020motion_attentive}) by a margin of 2.4\% in Mean-$\mathcal{F}$ and 1.0\% in Mean-$\mathcal{J}$, achieving the new SOTA performance. 
Notably, the proposed U-VOS model also outperforms the semi-supervised model (\eg, AGA~\cite{johnander2019generative}), even though it utilizes the first ground-truth mask to reference object location.

We also compare \ourmodel~against 13 SOTA V-SOD models. The non-binary saliency maps\footnote{Note that all compared maps in the V-SOD task, including ours, are non-binary.} are obtained from the standard benchmark~\cite{fan2019shifting}. 
This can be seen from~\tabref{tab:score_VOS}, our method consistently outperforms all other models since 2018 on all metrics. In particular, for the $S_\alpha$ and $F_\beta^{max}$ metrics, our method improves the performance by $\sim$2.0\% compared with the best AAAI'20-PCAS~\cite{gu2020pyramid} model.

\subsubsection{Evaluation on MCL dataset} 
This dataset has fuzzy object boundaries in the low-resolution frames due to fast object movements. 
Therefore, the overall performance is lower than on DAVIS$_{16}$. As shown in~\tabref{tab:score_VOS}, our method still stands out in these extreme circumstances, with a 3.0$\sim$8.0\% increase in all metrics compared with ICCV'19-RCR~\cite{yan2019semi} and CVPR'19-SSAV~\cite{fan2019shifting}. 

\subsubsection{Evaluation on FBMS dataset} 
This is one of the most popular VOS datasets with diverse attributes, such as interacting objects, dynamic backgrounds, and no per-frame annotation. As shown in \tabref{tab:score_VOS}, our model achieves competitive performance in terms of $\mathcal{M}$.
Further, compared to the previous best-performing SSAV~\cite{fan2019shifting}, it obtains improvements in other metrics, including $\mathcal{S}_{\alpha}$ (0.890 \textit{vs.} SSAV=0.879) and $E_{\xi}^{max}$ (0.935 \textit{vs.} SSAV=0.926), making it more suitable to the human visual system (HVS) as mentioned in~\cite{fan2017structure,fan2018enhanced}.

\begin{table*}[t!]
  \centering
  \footnotesize
  \renewcommand{\arraystretch}{1.0}
  \setlength\tabcolsep{6.0pt}
  \caption{Benchmarking results of 13 state-of-the-art V-SOD models on three subsets of DAVSOD$_{19}$~\cite{fan2019shifting}. `$\dag$' denotes that we generate non-binary saliency maps without CRF~\cite{krahenbuhl2011efficient} for a fair comparison. `N/A' means the results are not available. 
  }
  \label{tab:davsod_benchmark}
  \begin{tabular}{rr|cccc|cccc|cccc}
  \toprule
   & & \multicolumn{4}{c|}{DAVSOD$_{19}$-Easy35}
   &\multicolumn{4}{c|}{DAVSOD$_{19}$-Normal25}
   &\multicolumn{4}{c}{DAVSOD$_{19}$-Difficult20} \\
   \cline{3-14}
   &Model
   & \multicolumn{1}{c}{$S_\alpha\uparrow$}   &\multicolumn{1}{c}{$E_\xi^{max}\uparrow$} &\multicolumn{1}{c}{$F_\beta^{max}\uparrow$} & \multicolumn{1}{c|}{$\mathcal{M}\downarrow$}
   & \multicolumn{1}{c}{$S_\alpha\uparrow$}   &\multicolumn{1}{c}{$E_\xi^{max}\uparrow$} &\multicolumn{1}{c}{$F_\beta^{max}\uparrow$} & \multicolumn{1}{c|}{$\mathcal{M}\downarrow$}
   & \multicolumn{1}{c}{$S_\alpha\uparrow$}   &\multicolumn{1}{c}{$E_\xi^{max}\uparrow$} &\multicolumn{1}{c}{$F_\beta^{max}\uparrow$} & \multicolumn{1}{c}{$\mathcal{M}\downarrow$}\\
  \hline
  \multirow{5}{*}{\begin{sideways}2018\end{sideways}}
&MBN~\cite{li2018unsupervisedECCV} 
& 0.646 & 0.694 & 0.506 &0.109 & 0.597 & 0.665 & 0.436 &0.127 & 0.561 & 0.635 & 0.352 &0.140 \\
&FGRN~\cite{li2018flow}            
& 0.701 & 0.765 & 0.589 &0.095 & 0.638 & 0.700 & 0.468 &0.126 & 0.608 & 0.698 & 0.390 &0.131 \\
&SCNN~\cite{tang2018weakly}        
& 0.680 & 0.745 & 0.541 &0.127 & 0.589 & 0.685 & 0.425 &0.193 & 0.533 & 0.677 & 0.345 &0.234 \\
&DLVS~\cite{wang2017videoTIP}         
& 0.664 & 0.737 & 0.541 &0.129 & 0.599 & 0.670 & 0.416 &0.147 & 0.571 & 0.687 & 0.336 &0.128 \\
&SCOM~\cite{chen2018scom}          
& 0.603 & 0.669 & 0.473 &0.219 & N/A & N/A & N/A & N/A & N/A & N/A & N/A & N/A \\
\hline
\multirow{9}{*}{\begin{sideways}2019$\sim$2020\end{sideways}}
&RSE~\cite{xu2019video} 
& 0.577 & 0.663 & 0.417 &0.146 & 0.549 & 0.590 & 0.360 &0.170 & 0.555 & 0.644 & 0.306 &0.130 \\
&SRP~\cite{cong2019video}
& 0.575 & 0.655 & 0.453 &0.146 & 0.545 & 0.601 & 0.387 &0.169 & 0.555 & 0.682 & 0.341 &0.123 \\
&MESO~\cite{xu2019TMMvideo}
& 0.549 & 0.673 & 0.360 &0.159 & 0.542 & 0.597 & 0.354 &0.165 & 0.556 & 0.661 & 0.310 &0.127 \\
&LTSI~\cite{chen2019improved}
& 0.695 & 0.769 & 0.585 &0.106 & 0.658 & 0.723 & 0.499 &0.128 & 0.618 & 0.718 & 0.406 &0.112 \\
&SPD~\cite{li2019accurate}
& 0.626 & 0.685 & 0.500 &0.138 & 0.596 & 0.633 & 0.443 &0.171 & 0.574 & 0.688 & 0.345 &0.137 \\
&SSAV\cite{fan2019shifting}
& 0.755 & 0.806 & 0.659 &0.084 & 0.661 & 0.723 & 0.509 &0.117 & 0.619 & 0.696 & 0.399 &0.114 \\	
&RCR\cite{yan2019semi}
& 0.741 & 0.803 & 0.653 &0.087 & 0.674 & 0.729 & 0.533 &0.118 & 0.644 & \textbf{0.768} & 0.444 & \textbf{0.094} \\
&PCSA\cite{gu2020pyramid}
& 0.741 & 0.793 & 0.656 &0.086 & N/A & N/A & N/A & N/A & N/A & N/A & N/A & N/A \\
\hline
\rowcolor{mygray}
&\textbf{\ourmodel}$^\dag$
& \textbf{0.773} & \textbf{0.825} & \textbf{0.685} & \textbf{0.072} 
& \textbf{0.707} & \textbf{0.764} & \textbf{0.597} & \textbf{0.104} 
& \textbf{0.662} & 0.752 & \textbf{0.487} & 0.099 \\
\bottomrule
  \end{tabular}
\end{table*}

\subsubsection{Evaluation on SegTrack-V2 dataset} 
This is the earliest VOS dataset from the traditional era. Thus, only a limited number of deep U-VOS models have been tested on it. We only compare our \ourmodel~against the top-3 models: AAAI'20-PCAS~\cite{gu2020pyramid} ($S_\alpha$=0.866), ICCV'19-RCR~\cite{yan2019semi} ($S_\alpha$=0.842), and CVPR'19-SSAV~\cite{fan2019shifting} ($S_\alpha$=0.850). Our method achieves the best performance ($S_\alpha$=0.870).

\subsubsection{Evaluation on DAVSOD$_{19}$ dataset}

Recently published, DAVSOD$_{19}$~\cite{fan2019shifting} is the most challenging visual attention consistent V-SOD dataset with high-quality annotations and diverse attributes.
It contains diversified challenging scenarios due to the video sequences containing shifts in attention.
DAVSOD$_{19}$ is divided into three subsets, according to difficulty: DAVSOD$_{19}$-Easy-35 (35 clips), DAVSOD$_{19}$-Normal25 (25 clips), and DAVSOD$_{19}$-Difficult20 (20 clips). 
Note that, in the saliency field, non-binary maps are required for evaluation; thus, we only report the results of \ourmodel~without CRF post-processing in benchmarking the V-SOD task.
In this document, we adopt the four metrics mentioned in~\secref{sec:metric_vsod}, including $S_\alpha$, $E_{\xi}^{max}$, $F_\beta^{max}$, and $\mathcal{M}$.
For showing the robustness of \ourmodel, in~\tabref{tab:davsod_benchmark}, we also make the first effort to benchmark all 11 SOTA models since 2018, in terms of the three difficulty levels:

\begin{figure*}[t!]
  \centering
  \includegraphics[width=\linewidth]{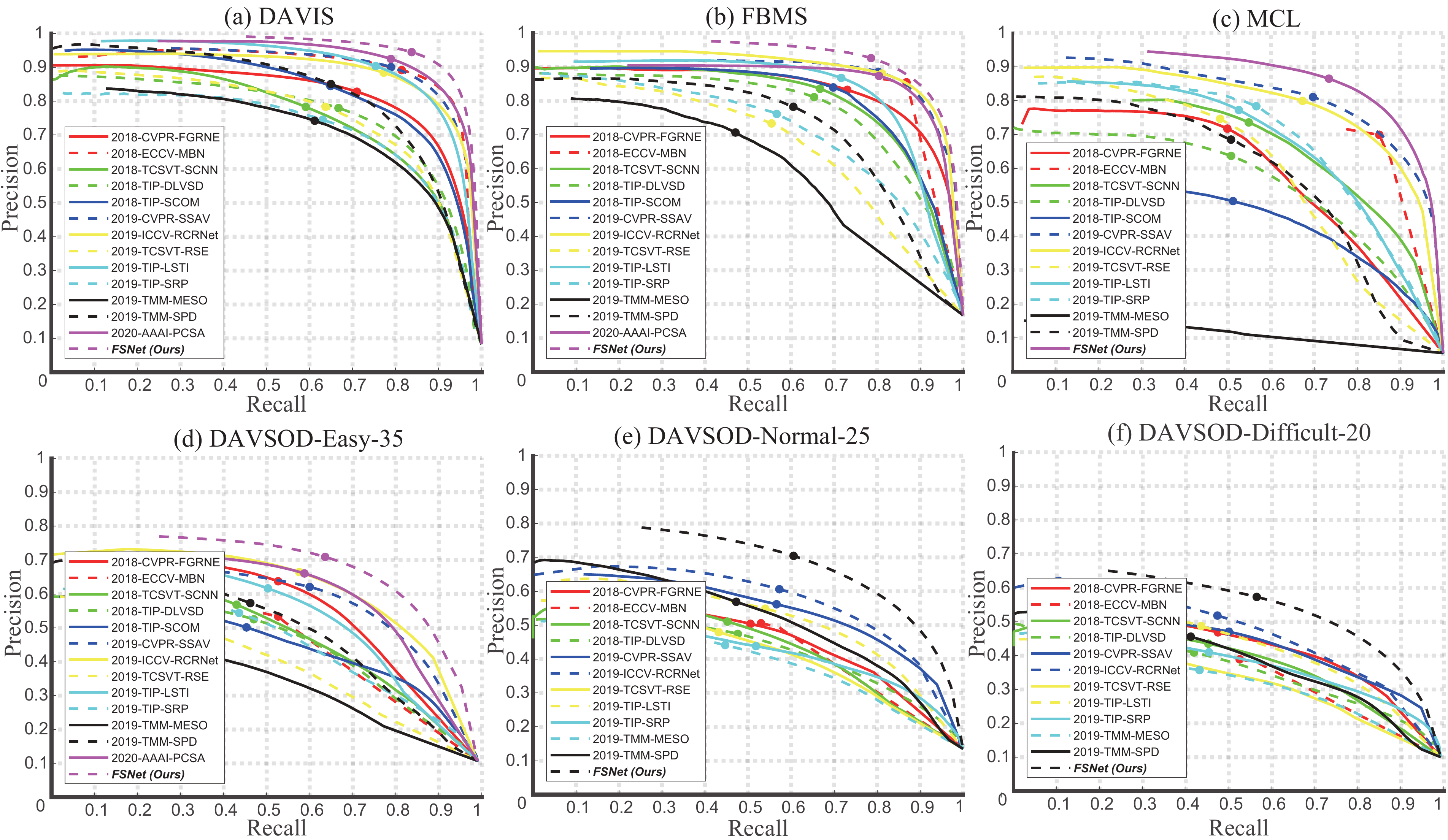}
  \caption{Precision-recall curves of SOTA V-SOD methods and the proposed \ourmodel~across six datasets. Zoom in for details and the best view in color for friendlier observation.}
  \label{fig:pr_curve}
\end{figure*}

\begin{itemize}
    \item \textbf{Easy35 subset:} Most of the video sequences are similar to those in the DAVIS$_{16}$ dataset, which also consists of a large number of single video objects. We see that 
    \ourmodel~outperforms all the reported algorithms across all metrics. 
    As shown in~\tabref{tab:davsod_benchmark}, compared with the recent method (PCSA), our model achieves large improvements of 3.2\% in terms of $S_{\alpha}$.

    \item \textbf{Normal25 subset:} Different from previous subsets, this one includes multiple moving salient objects. Thus, it is more difficult than traditional V-SOD datasets due to the attention shift phenomena~\cite{fan2019shifting}. As expected, \ourmodel~still obtains the best performance, with significant improvement, \eg, 6.4\% for $F_{\beta}^{max}$ metric.

    \item \textbf{Difficult20 subset:} This is the most challenging subset in existing V-SOD datasets since it contains a large number of attention shift sequences under cluttered scenarios.
    Therefore, from the results shown in~\tabref{tab:davsod_benchmark}, the performances of all the compared models decrease dramatically (\eg, $F_\beta^{max}\leq$ 0.5). 
    Even though our framework is not specifically designed for the V-SOD task, we still easily obtain the best performance in two metrics (\eg, $\mathcal{S}_{\alpha}$ and $F_\beta^{max}$).
    Different from the best two models, which utilize additional training data (\ie, RCR leverages pseudo-labels, SSAV utilizes the validation set), our model does not use any additional training data and still outperforms the SSAV model by 8.8\% ($F_{\beta}^{max}$), and achieves comparable performance to the second-best RCR (ICCV'19) model.
    These results are also supported by recent conclusions that ``human visual attention should be an underlying mechanism that drives U-VOS and V-SOD'' (TPAMI'20~\cite{wang2020paying}).
\end{itemize}

\subsubsection{PR Curve}\label{sec:pr_curve}
As shown in~\figref{fig:pr_curve}, we further investigate the precision-recall curves of different models on six V-SOD datasets, including DAVIS$_{16}$~\cite{perazzi2016benchmark}, MCL~\cite{kim2015spatiotemporal}, FBMS~\cite{ochs2013segmentation}, and DAVSOD$_{19}$~\cite{fan2019shifting} (\ie, Easy35, Normal25, and Difficult20).
Note that the higher and more to the right in the PR curve, the more accurate performance.
Even though existing SOTA methods have achieved significant progress in the V-SOD task on three typical benchmark datasets, we still obtain the best performance under all thresholds.
Besides, as a recent and challenging dataset, the overall performances on the three subsets of DAVSOD$_{19}$~\cite{fan2019shifting} are relatively poor. However, our \ourmodel~again achieves more satisfactory performance by large margins.

\begin{figure*}[t]
  \centering
  \includegraphics[width=\linewidth]{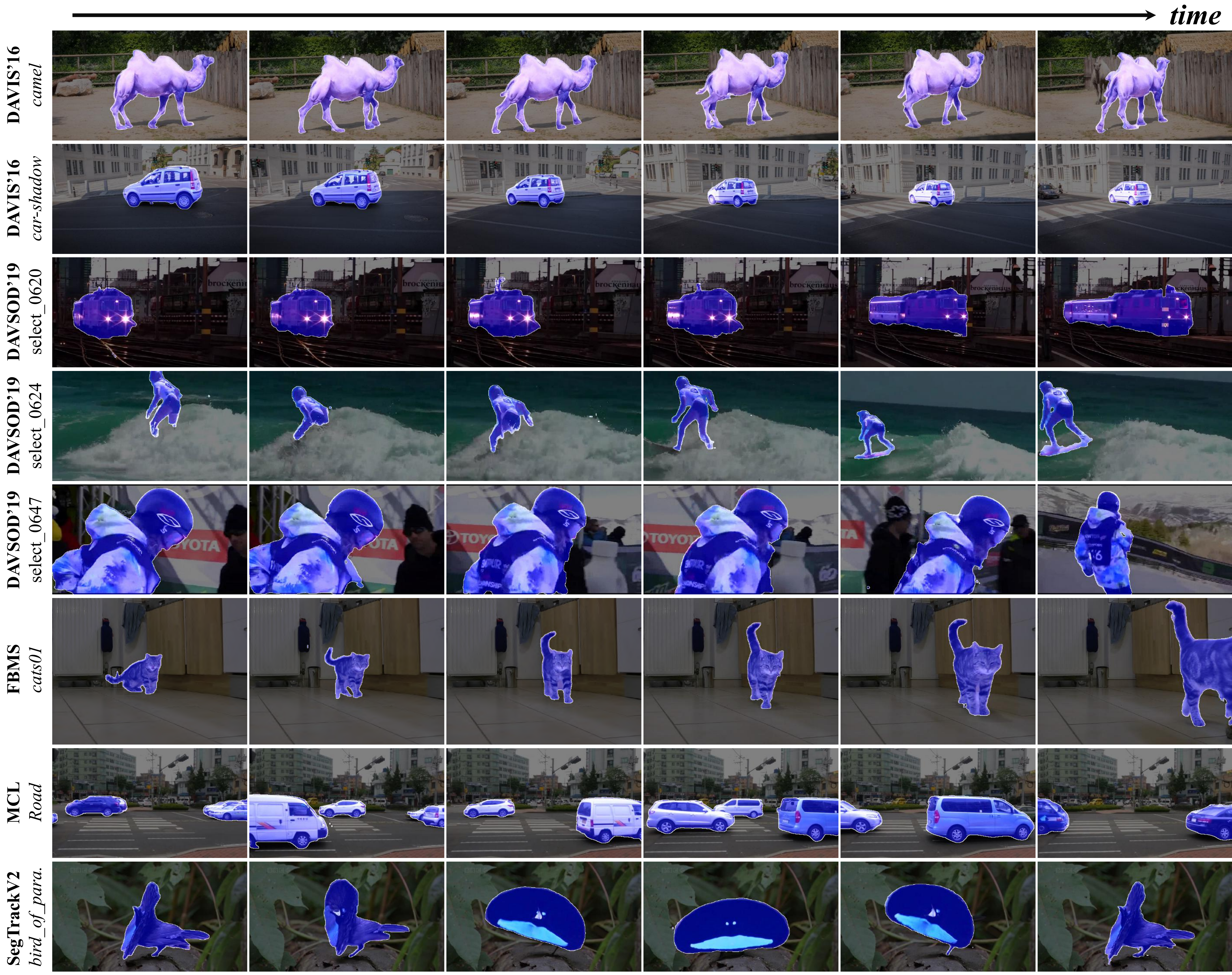}
  \caption{Qualitative results on five datasets, including  DAVIS$_{16}$~\cite{perazzi2016benchmark}, MCL~\cite{kim2015spatiotemporal},  FBMS~\cite{ochs2013segmentation}, SegTrack-V2~\cite{li2013video}, 
  and DAVSOD$_{19}$~\cite{fan2019shifting}. 
  }
\label{Fig:performance1}
\end{figure*}

\subsubsection{Qualitative Results}
Some qualitative results on the five datasets are shown in~\figref{Fig:performance1}, validating that our method achieves high-quality U-VOS and V-SOD results. 
As can be seen in the 1$^{st}$ row, the behind camel did not move, so it does not get noticed.
Interestingly, as our full-duplex strategy model considers both appearance and motion bidirectionally, it can automatically predict the dominated camel in the centre of the video instead of the camel behind.
A similar phenomenon is also presented in the 5$^{th}$ row, our method successfully detects dynamic skiers with the video clip rather than the static man in the background.
Overall, for these challenging situations, \eg, 
dynamic background (1$^{st}$ \& 5$^{th}$ rows), 
fast-motion (4$^{rd}$ row), 
out-of-view (6$^{rd}$ \& 7$^{nd}$ row),
occlusion (7$^{nd}$ row), and 
deformation (8$^{th}$ row), 
our model is able to infer the real target object(s) with fine-grained details. 
From this point of view, we demonstrate that \ourmodel~is a general framework for both U-VOS and V-SOD tasks.

\subsection{Ablation Study}

In this section, we conduct ablation studies to analyse our \ourmodel, including stimulus selection (\secref{sec:stimulus_selection}), effectiveness of RCAM (\secref{sec:effectiveness_of_RCAM}) and BPM (\secref{sec:effectiveness_of_bpm}), number of cascaded BPMs (\secref{sec:number_of_bpms}), and effectiveness of full-duplex strategy (\secref{sec:effectivess_of_FS}).

\subsubsection{Stimulus Selection}\label{sec:stimulus_selection}
We explore the influence of different stimuli (appearance only \emph{vs.} motion only) in our framework. We use only video frames or motion maps (using~\cite{ilg2017flownet}) to train the ResNet-50~\cite{he2016deep} backbone together with the proposed decoder block (see~\secref{sec:decoder}). 
As shown in~\tabref{tab:ablation_icam}, $Mo.$ performs slightly better than $App.$ in terms of $S_\alpha$ on DAVIS$_{16}$, which suggests that the ``optical flow'' setting can learn more visual cues than ``video frames''. 
Nevertheless, $App.$ outperforms $Mo.$ in $\mathcal{M}$ metric on MCL. 
This motivates us to explore how to use appearance and motion cues simultaneously effectively.

\subsubsection{Effectiveness of RCAM}\label{sec:effectiveness_of_RCAM}
To validate our RCAM (Rel.) effectiveness, we replace our fusion strategy with the vanilla fusion (Vanilla) using a concatenate operation followed by a convolutional layer to fuse two modalities. 
As expected (\tabref{tab:ablation_icam}), the proposed Rel. performs consistently better than the vanilla fusion strategy on both DAVIS$_{16}$ and MCL datasets.
We would like to point out that our RCAM has two important properties: 
\begin{itemize}
\item It enables mutual correction and attention.

\item It can alleviate error propagation within a network to an extent due to the mutual correction and bidirectional interaction.
\end{itemize}
%

\begin{table}[t!]
  \centering
  \scriptsize
  \renewcommand{\arraystretch}{1.2}
  \setlength\tabcolsep{1.8pt}
  \caption{Ablation studies (\secref{sec:stimulus_selection}, \secref{sec:effectiveness_of_RCAM}, \& \secref{sec:effectiveness_of_bpm}) for our components on DAVIS$_{16}$ and MCL. We set $N=4$ for BPM. 
  }\label{tab:ablation_icam}
  \begin{tabular}{r|cccc|cc|cc}
  \toprule
    &\multicolumn{4}{c|}{Component Settings} & \multicolumn{2}{c|}{DAVIS$_{16}$} &\multicolumn{2}{c}{MCL} \\
    \cline{2-9}
   & Appearance & Motion & RCAM & BPM 
   & $\mathcal{S}_{\alpha}\uparrow$ &$\mathcal{M}\downarrow$
   &$\mathcal{S}_{\alpha}\uparrow$ & $\mathcal{M}\downarrow$ \\
  \hline
  $App.$& \checkmark& & &                                  & 0.834 & 0.047 & 0.754 & 0.038 \\
  $Mo.$ & &\checkmark & &                                  & 0.858 & 0.039 & 0.763 & 0.053 \\
  Vanilla & \checkmark &\checkmark &  &                & 0.871 & 0.035 & 0.776 & 0.046 \\
  Rel.&\checkmark &\checkmark & \checkmark & & 0.900 & 0.025 & 0.833 & 0.031 \\
  Bi-Purf.& \checkmark &\checkmark &&\checkmark    & 0.904 & 0.026 & 0.855 & 0.023 \\
  \rowcolor{mygray}
  \hline
  \textbf{\ourmodel$^\dag$}&\checkmark &\checkmark &\checkmark &\checkmark & \textbf{0.920} & \textbf{0.020} & \textbf{0.864} & \textbf{0.023} \\
  \bottomrule
  \end{tabular}
\end{table}

\begin{table}[t!]
  \centering
  \scriptsize
  \renewcommand{\arraystretch}{1.2}
  \setlength\tabcolsep{4pt}
  \caption{Ablation study for the number ($N$) of BPMs on DAVIS$_{16}$~\cite{perazzi2016benchmark} and MCL~\cite{kim2015spatiotemporal}, focusing on parameter and FLOPs of BPMs, and runtime of~\ourmodel.
  }
  \label{tab:number_of_bpm}
  \begin{tabular}{r|c|c|c|cc|cc}
  \toprule
   & Param. & FLOPs & Runtime & \multicolumn{2}{c|}{DAVIS$_{16}$} &\multicolumn{2}{c}{MCL} \\
  \cline{5-8}
   & (M) & (G) & (s/frame) 
   & $\mathcal{S}_{\alpha}\uparrow$ & $\mathcal{M}\downarrow$
   & $\mathcal{S}_{\alpha}\uparrow$ & $\mathcal{M}\downarrow$ \\
  \hline
  $N=0$ & 0.000 & 0.000  & 0.03 & 0.900 &  0.025 & 0.833  & 0.031\\
  $N=2$ & 0.507 & 1.582 & 0.05 & 0.911 & 0.026 & 0.843 & 0.028 \\
  \rowcolor{mygray}
  $N=4$
  & 1.015 & 3.163 & 0.08 & \textbf{0.920} &  \textbf{0.020} & \textbf{0.864} & \textbf{0.023} \\
  $N=6$ & 1.522 & 4.745 & 0.10 & 0.918  & 0.023 & 0.863  & 0.023 \\
  $N=8$ & 2.030 & 6.327 & 0.13 & 0.920& 0.023 & 0.864 & 0.023 \\
  \bottomrule
  \end{tabular}
\end{table}

\subsubsection{Effectiveness of BPM}\label{sec:effectiveness_of_bpm}

To illustrate the effectiveness of the BPM (with $N=4$), we derive two different models: Rel. and \ourmodel, referring to the framework \textit{without} or \textit{with} BPM. We observe that the model with BPM gains 2.0$\sim$3.0\% than the one without BPM, according to the statistics in \tabref{tab:ablation_icam}. We attribute this improvement to BPM's introduction of an interlaced decremental connection, enabling it to fuse the different signals effectively.
Similarly, we remove the RCAM and derive another pair of settings (Vanilla \& Bi-Purf.) to test the robustness of our BPM. The results show that even using the bidirectional vanilla fusion strategy (Bi-Purf.) can still enhance the stability and generalization of the model. This benefits from the purification forward process and re-calibration backward process in the whole network.

\begin{table}[t!]
  \centering
  \scriptsize
  \renewcommand{\arraystretch}{1.2}
  \setlength\tabcolsep{1.2pt}
  \caption{Ablation study for the \textit{simplex} and \textit{full-duplex} strategies on DAVIS$_{16}$~\cite{perazzi2016benchmark} and MCL~\cite{kim2015spatiotemporal}. We set $N=4$ for BPM. 
  }
  \label{tab:ablation_bi_model}
  \begin{tabular}{c|c|c|cc|cc}
  \toprule
   &\multicolumn{2}{c|}{Direction Setting} & \multicolumn{2}{c|}{DAVIS$_{16}$} &\multicolumn{2}{c}{MCL} \\
   \cline{2-7}
   & \multicolumn{1}{c|}{RCAM} & \multicolumn{1}{c|}{BPM} 
   & $\mathcal{S}_{\alpha}\uparrow$  & $\mathcal{M}\downarrow$
   & $\mathcal{S}_{\alpha}\uparrow$ & $\mathcal{M}\downarrow$\\
  \hline
  \multirow{4}{*}{\begin{sideways} simplex \end{sideways}}
  &$App. \Rightarrow Mo.$     &$(App.+Mo.) \Rightarrow Mo.$ 
  & 0.896 & 0.026 & 0.816 & 0.038  \\
  &$App. \Rightarrow Mo.$     &$(App.+Mo.) \Leftarrow Mo.$  
  & 0.902 & 0.025 & 0.832 & 0.031   \\
  &$App. \Leftarrow Mo.$      &$(App.+Mo.) \Rightarrow Mo.$ 
  & 0.891 & 0.029 & 0.806 & 0.039  \\
  &$App. \Leftarrow Mo.$      &$(App.+Mo.) \Leftarrow Mo.$ 
  & 0.897 & 0.028 & 0.840 & 0.028  \\
  \hline
  self-purf. &$App. \Leftrightarrow Mo.$ &$(App.+Mo.) \nLeftrightarrow Mo.$ & 0.899 & 0.026 & 0.854 & 0.023\\
  \hline
  \rowcolor{mygray}
  full-dup. & $App. \Leftrightarrow Mo.$ & $(App.+Mo.) \Leftrightarrow Mo.$ 
  & \textbf{0.920} & \textbf{0.020} & \textbf{0.864} & \textbf{0.023} \\
  \bottomrule
  \end{tabular}
\end{table}

\subsubsection{Number of Cascaded BPMs}\label{sec:number_of_bpms}
Naturally, more cascaded BPMs should lead to better boost performance. This is investigated and the evaluation results are shown in \tabref{tab:number_of_bpm}, where $N=\{0,2,4,6,8\}$.
Note that $N=0$ means that \textbf{NO} BPM is used.
%
%
Clearly, as can be seen from~\figref{fig:front_figure_2} (red star), we compare four variants of our~\ourmodel, including $N$=0 (Mean-$\mathcal{J}$=76.4, Mean-$\mathcal{F}$=76.8), $N$=2 (Mean-$\mathcal{J}$=80.4, Mean-$\mathcal{F}$=81.4), $N$=4 (Mean-$\mathcal{J}$=82.1, Mean-$\mathcal{F}$=83.3), and $N$=4, CRF (Mean-$\mathcal{J}$=83.4, Mean-$\mathcal{F}$=83.1). 
It demonstrates that more BPMs leads to better results, but the performance reaches saturation after $N=4$. 
Further, too many BPMs (\ie, $N>4$) will cause high model-complexity and increase the over-fitting risk. As a trade-off, we use $N=4$ throughout our experiments.

\subsubsection{Effectiveness of Full-Duplex Strategy}\label{sec:effectivess_of_FS}
To investigate the effectiveness of the RCAM and BPM modules with the full-duplex strategy, we study two unidirectional (\ie, simplex strategy in \figref{Fig:framework_RCAM} \& \figref{Fig:framework_bpm}) variants of our model. 
In \tabref{tab:ablation_bi_model}, the symbols $\Rightarrow$, $\Leftarrow$, and $\Leftrightarrow$ indicate the feature transmission directions in the designed RCAM 
or BPM.
Specifically, $App. \Leftarrow Mo.$ indicates that the attention vector in the optical flow branch weights the features in the appearance branch and vice versa. $(App. + Mo.) \Leftarrow Mo.$ indicates that motion cues are used to guide the fused features extracted from both appearance and motion. 
The comparison results show that our elaborately designed modules (RCAM and BPM) jointly cooperate in a full-duplex fashion and outperform all simplex (\textit{unidirectional}) settings.

\subsection{Further Discussion}\label{sec:discussion}
\subsubsection{Prediction Selection}\label{sec:prediction_selection}
Which is the final prediction, $\mathbf{S}_{A}^{t}$ or $\mathbf{S}_{M}^{t}$?
As mentioned in~\secref{sec:training}, we choose $\mathbf{S}_{A}^{t}$ as our final segmentation result instead of $\mathbf{S}_{M}^{t}$. 
The major reasons for doing so can be summarized as follows:
\begin{itemize}
    \item We employ the auxiliary supervision for the motion-based branch to learn more motion patterns inspired by~\cite{tokmakov2017learning2}.
    \item More informative appearance and motion cues are contained in another branch at the phase of bidirectional purification.
\end{itemize}

\begin{table}[htbp]
  \centering
  \scriptsize
  \renewcommand{\arraystretch}{1.2}
  \setlength\tabcolsep{8.5pt}
  \caption{Ablation study (\secref{sec:prediction_selection}) for the choice of final segmentation result on DAVIS$_{16}$~\cite{perazzi2016benchmark} and MCL~\cite{kim2015spatiotemporal} dataset.
  }
  \label{tab:choice_of_result}
  \begin{tabular}{l|cc|cc}
  \toprule
   & \multicolumn{2}{c|}{DAVIS$_{16}$} &\multicolumn{2}{c}{MCL} \\
   \cline{2-5}
   & $\mathcal{S}_{\alpha}\uparrow$ & $\mathcal{M}\downarrow$
   & $\mathcal{S}_{\alpha}\uparrow$ & $\mathcal{M}\downarrow$ \\
  \hline
  \hline
  (a) $\mathbf{S}_{M}^{t}$ as result 
  & 0.920 & 0.022 & 0.862 & 0.024 \\
  (b) $(\mathbf{S}_{A}^{t}+\mathbf{S}_{M}^{t})/2$ as result 
  & 0.920 & 0.022 & 0.863 & 0.023\\
  \hline
  \rowcolor{mygray}
  (c) $\mathbf{S}_{A}^{t}$ as result (Ours) 
  & 0.920 & 0.022 & 0.864 & 0.023 \\
  \toprule
  \end{tabular}
\end{table}

As shown in~\tabref{tab:choice_of_result}, three experiments are conducted to verify our assumption: 
(a) choosing $\mathbf{S}_{M}^{t}$ as the final result, 
(b) choosing $(\mathbf{S}_{A}^{t}+\mathbf{S}_{M}^{t})/2$ as the final result, and 
(c) choosing $\mathbf{S}_{A}^{t}$ as the final result (Ours). 
As can be seen in~\tabref{tab:choice_of_result}, all three choices achieve very similar results, while $\mathbf{S}_{A}^{t}$ performs slightly better than the other two.
Besides, considering the reduction of unnecessary computational cost, we choose $\mathbf{S}_{A}^{t}$ as our final result for comparison with other methods.

\subsubsection{Effectiveness of CRF}\label{sec:crf_post_process}
\begin{figure}[htbp]
  \centering
  \begin{overpic}[width=0.8\linewidth]{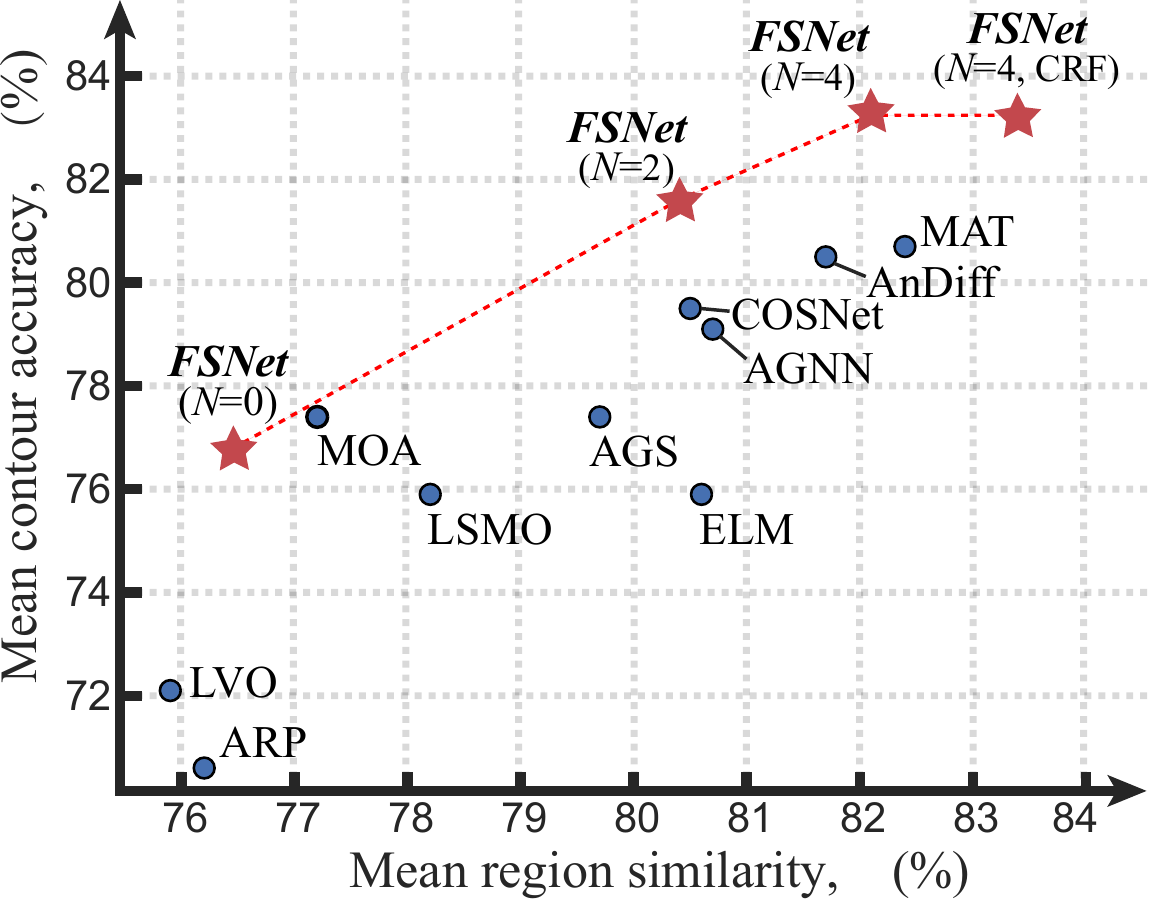}
    \put(74, 1){\small $\mathcal{J}$}
    \put(3, 64){\begin{rotate}{90}{\small $\mathcal{F}$}\end{rotate}}
  \end{overpic}
  \caption{
  Mean contour accuracy ($\mathcal{F}$)~\textit{vs.}~mean region similarity ($\mathcal{J}$) scores on DAVIS$_{16}$ dataset~\cite{perazzi2016benchmark}. 
  Circles indicate U-VOS methods.
  Four variants are shown in \textit{\textbf{bold-italic}}, in which `$N$' indicates the number of bidirectional purification modules (BPM) and `CRF' means that using CRF~\cite{krahenbuhl2011efficient} post-processing technique.
  Compared with the best unsupervised VOS model (\ie, MAT~\cite{zhou2020motion_attentive} also with CRF),
  the proposed method \ourmodel~($N$=4, CRF) achieves the new SOTA by a large margin.
  }
  \label{fig:front_figure_2}
\end{figure}

From \figref{fig:front_figure_2} we can see that our~\ourmodel~without CRF post-processing technique, \ie, \ourmodel~($N$=4), still outperforms the best model AAAI'20-MAT in terms of Mean-$\mathcal{F}$ metric.
This means that our initial method (\ie, \ourmodel~without CRF) can distinguish hard samples around the object boundaries without post-processing techniques.
When equipped with the CRF post-processing technique~\cite{krahenbuhl2011efficient}, our \ourmodel~($N$=4, CRF) achieves the best performance in terms of both Mean-$\mathcal{J}$ and Mean-$\mathcal{F}$ metrics.

\begin{figure}[t!]
  \centering
  \includegraphics[width=\linewidth]{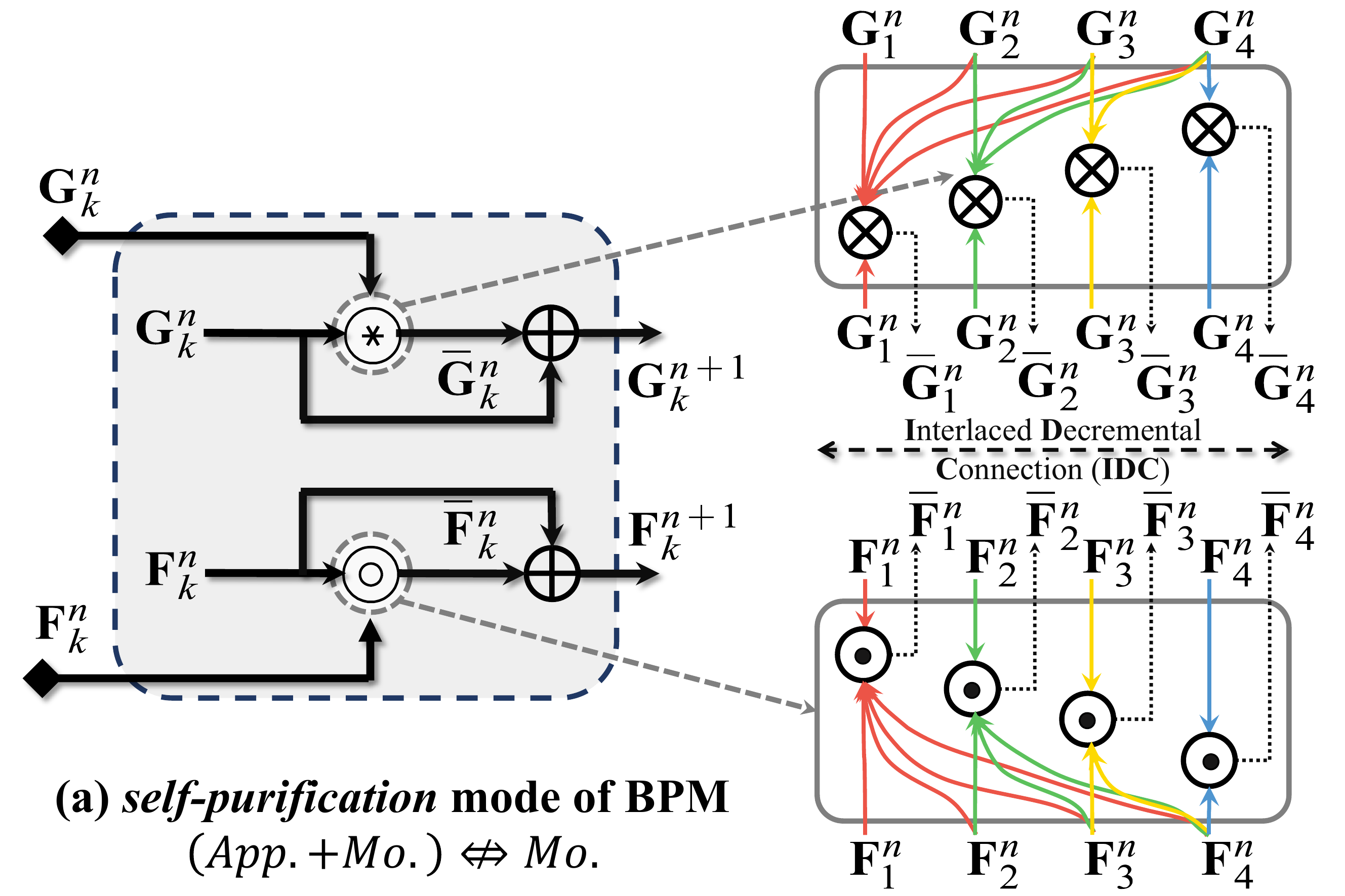}
  \caption{Illustration of self-purification strategy (a) and the proposed bidirectional purification strategy (b). Note that sub-figure (b) is same as the \figref{Fig:framework_bpm}~(c) for convenient comparison. Note that $\oplus$, $\otimes$, and $\odot$ denote element-wise addition, multiplication, and concatenation, respectively.
  }\label{fig:self_purf}
\end{figure}

\subsubsection{Training Effectiveness with Less Data}\label{sec:scale_of_dataset}
As shown in~\figref{fig:front_figure_2}, the proposed method, \ie, \ourmodel~($N$=4, CRF), surpasses the best U-VOS model MAT~\cite{zhou2020motion_attentive} (also with CRF), while our~\ourmodel~with less labelled data in the training phase (\ie, Ours-13K \textit{vs.} MAT-16K). 
Besides, we also observe that the recently proposed method 3DC-Seg~\cite{bmvc2020making}, based on a 3D convolutional network, can achieve the new state-of-the-art (Mean-$\mathcal{J}$=84.3, Mean-$\mathcal{F}$=84.7), while relies on a massive amount of labelled training samples as expert knowledge in the fine-tuning phase, including 158K images (\ie, COCO~\cite{lin2014microsoft} + YouTube-VOS~\cite{xu2018youtube} + DAVIS$_{16}$~\cite{perazzi2016benchmark}).
%
%
It requires about ten times more training data than the best model MAT~\cite{zhou2020motion_attentive} (16K images) in the fine-tuning phase. 
%
%
Thus, it demonstrates the efficient training process in our pipeline.
%

\subsubsection{Self-Purification Strategy in BPM}\label{sec:self_purf}
We provide more details on the different variants mentioned in~\secref{sec:effectivess_of_FS}
including $(App. + Mo.) \Leftarrow Mo.$, $(App. + Mo.) \Rightarrow Mo.$ and $(App.+Mo.) \Leftrightarrow Mo.$ in BPM.
The implementation of $App. \Leftarrow Mo.$ and $App. \Rightarrow Mo.$ in RCAM are illustrated in~\figref{Fig:framework_RCAM}~(a)~\&~(b), whereas the structure implementation of $(App. + Mo.) \Leftarrow Mo.$ \& $(App. + Mo.) \Rightarrow Mo.$ in BPM are illustrated in~\figref{Fig:framework_bpm}~(a)~\&~(b).
Here, note that all of these variants indicate unidirectional refinement, in contrast to the proposed bi-directional schemes.

Last but not least, to validate that the gains of bi-directional schemes in practice \textbf{DO COME FROM} the bi-directional procedure and not more complex model structures, we implement another variant using the same complex structures but without any branch interactions before the decoding stage.
This is done by exchanging the places of $\mathbf{G}_{k}^n$ and $\mathbf{F}_{k}^n$ as illustrated in Fig.~\ref{fig:self_purf} (b), leading to a kind of ``self-purification'' strategy. Symbol ``$\nLeftrightarrow$'' in Fig.~\ref{fig:self_purf} (a) means that there is \textbf{NO} interaction between the two branches, \ie, there is only interaction within itself.
Comparisons of the uni-/bidirectional strategies are shown in~\tabref{tab:ablation_bi_model}. The comparison results show that
our elaborately designed modules (\ie, RCAM and BPM) jointly cooperate in a bidirectional manner and outperform all unidirectional settings.
Besides, our bidirectional purification scheme (\ie, `full-dup.'~in~\tabref{tab:ablation_bi_model}) also achieves very notable improvement (2.1\% and 1.0\% gains in $S_\alpha$ on DAVIS$_{16}$~\cite{perazzi2016benchmark} and MCL~\cite{kim2015spatiotemporal}, respectively) against the ``self-purification'' variant (\ie, `self-purf.' in~\tabref{tab:ablation_bi_model}), which has a similar complex structure, further validating the benefit of the bidirectional behavior claimed in this study.
%

\subsubsection{Relation Between RCAM and BPM}\label{sec:relation_RCAM_BPM}
The two introduced modules, \ie, RCAM and BPM, focus on using appearance and motion features while ensuring the information flow between them.
They can work collaboratively under the mutual restraint of our full-duplex strategy, but they cannot be substituted for one another.
This is due to the RCAM transmits the features at each level in a \textit{point-to-point} manner (\eg, $\mathcal{X}_1 \rightarrow \mathcal{Y}_1 $), and thus, it fits with the progressive feature extraction in the encoder. 
The BPM, on the other hand, broadcasts high-level features to low-level features via an interlaced decremental connection in a \textit{set-to-point} manner (\eg,  $ \{ \mathbf{F}_2^n, \mathbf{F}_3^n, \mathbf{F}_4^n\} \rightarrow \mathbf{G}_2^n$), which is more suitable for the multi-level feature interaction in the decoder.

\section{Conclusion}
In this paper, we present a simple yet efficient framework, termed full-duplex strategy network (\ourmodel), that fully leverages the mutual constraints of appearance and motion cues to address the video object segmentation problem. 
It consists of two core modules: 
the relational cross-attention module (RCAM) in the encoding stage and the efficient bidirectional purification module (BPM) in the decoding stage.
The former one is used to abstract features from a dual-modality, while the latter is utilized to re-calibrate inconsistant features step-by-step.
%
%
We thoroughly validate functional modules of our architecture via extensive experiments, leading to several interesting findings.
Finally, \ourmodel~acts as a unified solution that significantly advances SOTA models for both U-VOS and V-SOD tasks.
In the future, we may extend our scheme to learn short-term and long-term information in an efficient Transformer-like framework \cite{wang2021pyramid,zhuge2021kaleido} to further boost the accurarcy.

\bibliographystyle{CVM}
{\normalsize  \bibliography{CVM2021-VOS}}

\end{document}